\RequirePackage[l2tabu,orthodox]{nag}

\documentclass[headsepline,footsepline,footinclude=false,oneside,fontsize=11pt,paper=a4,listof=totoc,bibliography=totoc]{scrbook} % one-sided

\PassOptionsToPackage{table,svgnames,dvipsnames}{xcolor}

\usepackage[utf8]{inputenc}
\usepackage[T1]{fontenc}
\usepackage[sc]{mathpazo}
\usepackage[ngerman,american]{babel}
\usepackage[autostyle]{csquotes}
\usepackage[natbib=true, 
            url=false, 
            style=authoryear,
            maxcitenames=2,
            maxbibnames=99,          
            date=year,
            backref=true
            ]{biblatex}
\usepackage{todonotes}
\usepackage{graphicx}
\usepackage{scrhack} % necessary for listings package
\usepackage{listings}
\usepackage{lstautogobble}
\usepackage{multirow}
\usepackage{tikz}
\usepackage{pgfplots}
\usepackage{pgfplotstable}
\usepackage{subcaption}
\usepackage{booktabs}
\usepackage[final]{microtype}
\usepackage{caption}
\usepackage[printonlyused]{acronym}
\AtBeginDocument{%
	\hypersetup{
		pdftitle=\getTitle,
		pdfauthor=\getAuthor,
	}
}
\usepackage{ifthen}
\usepackage{amsmath}
\usepackage{amssymb}
\usepackage{amsthm}
\usepackage{wrapfig}
\usepackage{amsfonts}
\usepackage[normalem]{ulem} % for \sout

\usepackage{xcolor}
\definecolor{standardblue}{rgb}{0.149,0.580,0.871}
\usepackage[colorlinks,citecolor=standardblue]{hyperref}

\makeatletter
\if@twoside
\else
\fi
\makeatother

\addto\extrasamerican{

}

\addto\extrasngerman{
	
}

\ifthenelse{\equal{\detokenize{dark}}{\jobname}}{%
  \newcommand{\bg}{black} % background
  \newcommand{\fg}{white} % foreground
  \usepackage[pagecolor=\bg]{pagecolor}
  \color{\fg}
}{%
  \newcommand{\bg}{white} % background
  \newcommand{\fg}{black} % foreground
}

\setkomafont{disposition}{\normalfont\bfseries} % use serif font for headings
\BeforeTOCHead[toc]{{\cleardoublepage\pdfbookmark[0]{\contentsname}{toc}}}

\definecolor{TUMBlue}{HTML}{0065BD}
\definecolor{TUMSecondaryBlue}{HTML}{005293}
\definecolor{TUMSecondaryBlue2}{HTML}{003359}
\definecolor{TUMBlack}{HTML}{000000}
\definecolor{TUMWhite}{HTML}{FFFFFF}
\definecolor{TUMDarkGray}{HTML}{333333}
\definecolor{TUMGray}{HTML}{808080}
\definecolor{TUMLightGray}{HTML}{CCCCC6}
\definecolor{TUMAccentGray}{HTML}{DAD7CB}
\definecolor{TUMAccentOrange}{HTML}{E37222}
\definecolor{TUMAccentGreen}{HTML}{A2AD00}
\definecolor{TUMAccentLightBlue}{HTML}{98C6EA}
\definecolor{TUMAccentBlue}{HTML}{64A0C8}

\pgfplotsset{compat=newest}
\pgfplotsset{
  cycle list={TUMBlue\\TUMAccentOrange\\TUMAccentGreen\\TUMSecondaryBlue2\\TUMDarkGray\\},
}

\newcommand{\R}{\mathbb{R}}
\newcommand{\calD}{\mathcal{D}}

\newcommand{\bbE}{\mathbb{E}}

\newcommand{\U}{\mathcal{U}}
\newcommand{\N}{\mathcal{N}}
\newcommand{\divergence}{\operatorname{div}}
\newcommand{\trace}{\operatorname{Tr}}
\newcommand{\Cov}{\operatorname{Cov}}
\newcommand{\W}{\mathbf{W}}
\newcommand{\bP}{\mathbf{P}}
\newcommand{\bb}{\mathbf{b}}

\newcommand{\Lfm}{\mathcal{L}_{\text{FM}}(\theta)}
\newcommand{\Lcfm}{\mathcal{L}_{\text{CFM}}(\theta)}

\newcommand*{\getUniversity}{Technische Universität München}
\newcommand*{\getFaculty}{Informatics}
\newcommand*{\getDegree}{Informatics}
\newcommand*{\getSchool}{Computation, Information and Technology}
\newcommand*{\getTitle}{Geometric Flow Models over Neural Network Weights}

\newcommand*{\getAuthor}{Ege Erdogan}
\newcommand*{\getDoctype}{Master's Thesis}

\newcommand*{\getSubmissionDate}{08.01.2025}
\newcommand*{\getSubmissionLocation}{Munich}

\begin{document}

% Set page numbering to avoid "destination with the same identifier has been already used" warning for cover page.
% (see https://en.wikibooks.org/wiki/LaTeX/Hyperlinks#Problems_with_Links_and_Pages).
\pagenumbering{alph}
\begin{titlepage}
  % HACK for two-sided documents: ignore binding correction for cover page.
  % Adapted from Markus Kohm's KOMA-Script titlepage=firstiscover handling.
  % See http://mirrors.ctan.org/macros/latex/contrib/koma-script/scrkernel-title.dtx,
  % \maketitle macro.
  \oddsidemargin=\evensidemargin\relax
  \textwidth=\dimexpr\paperwidth-2\evensidemargin-2in\relax
  \hsize=\textwidth\relax

  \centering

  \IfFileExists{logos/tum-\fg.pdf}{%
    \includegraphics[height=20mm]{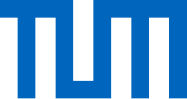}
  }{%
    \vspace*{20mm}
  }

  \vspace{5mm}
  {\huge\MakeUppercase{School of \getSchool{} --- \getFaculty{}} \par}

  \vspace{5mm}
  {\large\MakeUppercase{\getUniversity{}} \par}

  \vspace{15mm}
  {\Large \getDoctype{} in \getDegree{} \par}

  \vspace{10mm}
  {\huge\bfseries \getTitle{} \par}

  \vspace{10mm}
  {\LARGE \getAuthor{}}

  \vspace{10mm}
  % {\large \textbf{Supervisors:} David Rügamer, Bastian Rieck}
  {\large \textbf{Supervisors} \\ David Rügamer (LMU Munich) \\
   Bastian Grossenbacher Rieck (University of Fribourg)}

  \IfFileExists{logos/faculty-\fg.pdf}{%
    \vfill{}
    \includegraphics[height=20mm]{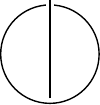}
  }{}
\end{titlepage}

\frontmatter{}

\thispagestyle{empty}
\vspace*{0.8\textheight}
\noindent
I confirm that this \MakeLowercase{\getDoctype{}} is my own work and I have documented all sources and material used.

\vspace{15mm}
\noindent
\getSubmissionLocation{}, \getSubmissionDate{} \hspace{\fill} \getAuthor{}

\cleardoublepage{}

\chapter{\abstractname}

Deep generative models such as flow and diffusion models have proven to be effective in modeling high-dimensional and complex data types such as videos or proteins, and this has motivated their use in different data modalities, such as neural network weights. A generative model of neural network weights would be useful for a diverse set of applications, such as Bayesian deep learning, learned optimization, and transfer learning. However, the existing work on weight-space generative models often ignores the symmetries of neural network weights, or only takes into account a subset of them. Modeling those symmetries, such as permutation symmetries between subsequent layers in an MLP, the filters in a convolutional network, or scaling symmetries arising with the use of non-linear activations, holds the potential to make weight-space generative modeling more efficient by effectively reducing the dimensionality of the problem. 

In this light, we aim to design generative models in weight-space that more comprehensively respect the symmetries of neural network weights. We build on recent work on generative modeling with flow matching, and weight-space graph neural networks to design three different weight-space flows. Each of our flows takes a different approach to modeling the geometry of neural network weights, and thus allows us to explore the design space of weight-space flows in a principled way. Our results confirm that modeling the geometry of neural networks more faithfully leads to more effective flow models that can generalize to different tasks and architectures, and we show that while our flows obtain competitive performance with orders of magnitude fewer parameters than previous work, they can be further improved by scaling them up. We conclude by listing potential directions for future work on weight-space generative models.

Our implementation is available at \url{https://github.com/ege-erdogan/weightflow}.
\microtypesetup{protrusion=false}
\tableofcontents{}
\microtypesetup{protrusion=true}

\mainmatter{}

% Background
% •⁠  ⁠bayesian dl
% •⁠  ⁠⁠diffusion/fm
% •⁠  ⁠⁠geometry of loss landscapes
% •⁠  ⁠⁠graph neural net and metanets
% Method/ design choices
% •⁠  ⁠fm design choices
% Results
% Discussion
% Future work

% \input{notation.tex}

% !TeX root = ../main.tex
% Add the above to each chapter to make compiling the PDF easier in some editors.

\begin{figure}[t!]
    \centering
    \includegraphics[width=\linewidth]{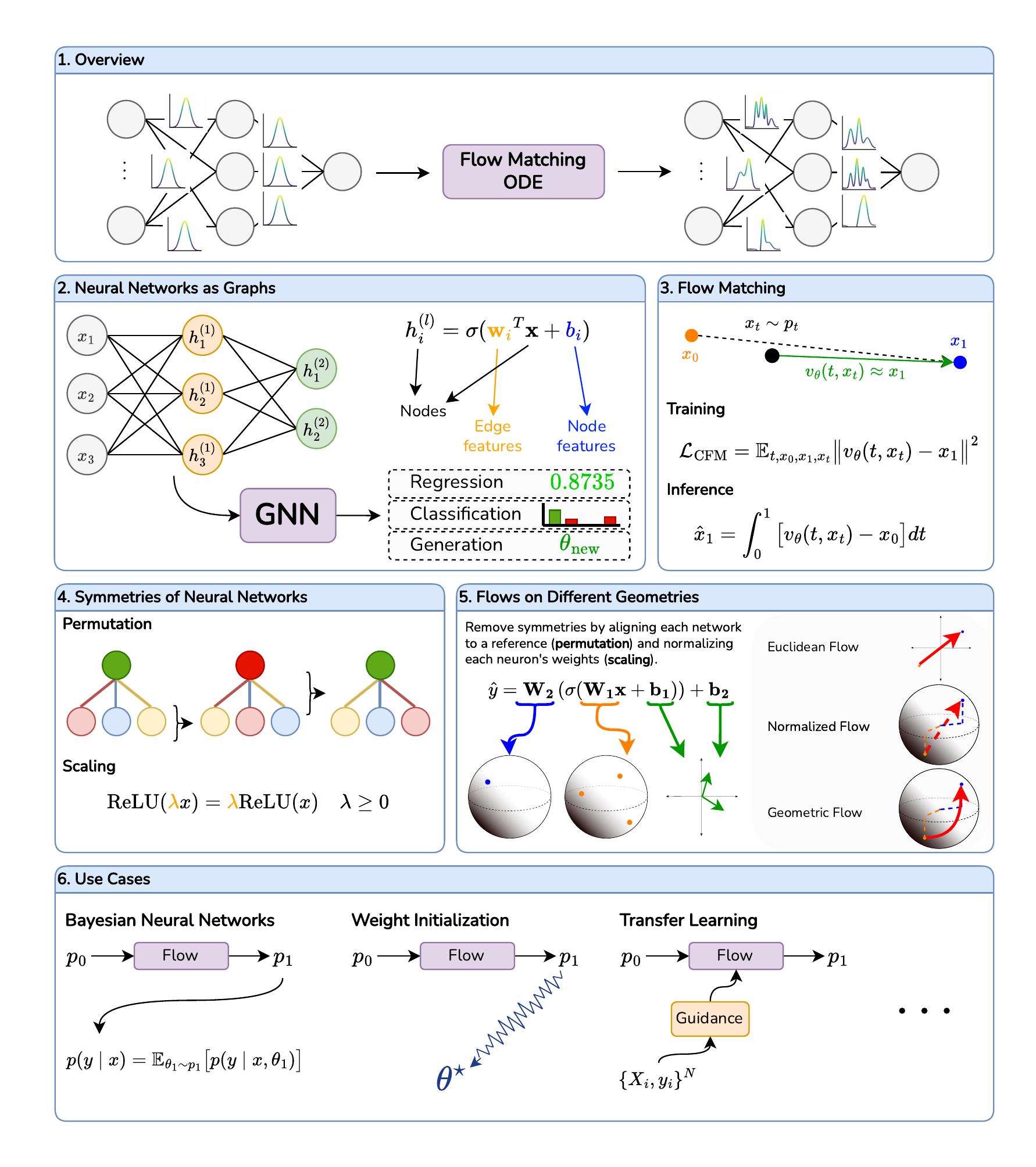}
    \caption{\label{fig:main}\textbf{Overview of our weight-space flow.} We aim to learn a flow in weight-space \textit{(1)}, processing neural networks with GNNs \textit{(2)}, using flow matching \textit{(3)} and taking into account the symmetries of neural network weights \textit{(4)}. We propose three different flows \textit{(5)} and potential use cases include Bayesian neural networks, learned weight initialization, or transfer learning \textit{(6)}.}
\end{figure}

\chapter{Introduction}\label{chapter:introduction}

Deep generative models such as diffusion and flow models that have led to significant developments in image generation \citep{esserScalingRectifiedFlow2024b} and biological applications such as protein structure prediction \citep{abramsonAccurateStructurePrediction2024} have also been applied to neural network weights \citep{peeblesLearningLearnGenerative2022,schurholtHyperRepresentationsLearningPopulations2024a}. However, the existing works do not take into account the geometry of neural networks arising from permutation and scaling symmetries, or only consider the permutation symmetries. However, modeling the symmetries of data often leads to improved performance and more data-efficient training \citep{brehmerDoesEquivarianceMatter2024}. We attempt to fill this gap by building flow models that learn a vector field to transport a prior over neural network to the posterior for a specific task through the flow matching framework \citep{lipmanFlowMatchingGuide2024}, processing neural network weights with permutation-invariant graph neural networks \citep{kofinasGraphNeuralNetworks2024,limGraphMetanetworksProcessing2023} and propose three candidate designs differing on how they handle the underlying geometry. Figure \ref{fig:main} gives an overview of our approach, and we summarize the main points in this Introduction. 

\section{Motivation}

\subsubsection*{Weight-Space Generative Models as Learned Optimizers}

A generative model of neural network weights is in essence a learned optimizer, and learned optimizers constitute an active research area in their own right \citep{hospedalesMetaLearningNeuralNetworks2022}. However, unlike typical learned optimizers that are often trained by unrolling the gradient-based optimization of neural networks \citep{finnModelAgnosticMetaLearningFast2017}, generative models are trained directly with data (trained neural network weights), using the outputs of neural network training. This implies weight-space generative models can be improved by utilizing public datasets of neural network checkpoints such as \citep{peeblesLearningLearnGenerative2022,schurholtModelZoosDataset2022}.

Inference with weight-space generative models is also fundamentally different than gradient-based optimization. Rather than predict the task gradients or output optimization steps, a generative model can output trained weights in ``one-shot.'' Furthermore, they open the way for applications of tools from the modern generative models literature to navigate the space of neural network weights. 

\subsubsection*{Weight-Space Generative Models as Probabilistic Models}

The fundamental units of modern generative models such as diffusion and flow models are probability distributions, that is they learn to transport one probability distribution such as a Normal distribution to another one such as the distribution of certain images. Weight-space generative models therefore also model a complex probability distribution over neural network weights. They can both efficiently sample from it, and evaluate the likelihood of arbitrary points under this distribution. 

Being able to both sample from and evaluate the likelihood function of a complex distribution puts generative models in a unique place among similar methods. The likelihood function of variational approximations of neural network posteriors can also be evaluated, but they model considerably simpler distributions, while Markov chain Monte Carlo methods can sample from more complex multi-modal distributions but do not allow likelihood computations. Nevertheless, this power of generative models comes with increased computational demands during training and the need for careful architecture design.

\subsubsection*{Generative Models, Geometry, and Neural Network Weights}

Geometric generative models that model the geometry of their domain by invariance/equivariance constraints \citep{kleinEquivariantFlowMatching2023a}, parametrizing their data points on manifolds \citep{chenRiemannianFlowMatching2023}, or learning a manifold structure from data points alone \citep{kapusniakMetricFlowMatching2024} have been applied to a wide range problems from climate modeling \citep{bodnarAuroraFoundationModel2024} to protein design \citep{boseSE3StochasticFlowMatching2024}. By baking in the correct inductive biases from the start rather than expecting a model to learn them from scratch or through data augmentation, such geometric models make training more data-efficient. 

It is thus a logical first-step, especially for high-dimensional and multi-modal distributions such as neural network posteriors, to model the geometric structure of the data as accurately as possible. In particular for deep neural networks, their non-identifiability - there being multiple parametrizations of the same function, is a key reason behind this multi-modality, and it is hypothesized that the modes in a neural network posterior are actually linearly connected up to function-preserving permutations \citep{entezariRolePermutationInvariance2022}. Together, these considerations motivate that weight-space learning tasks can be considerably simplified through appropriate geometric modeling choices.

\section{Overview}

\subsubsection{Symmetries of Neural Networks (Chapter \ref{section:geometry_of_nns})}

Neural network weights possess various symmetries, i.e. transformations of the network parameters that leave the function it is computing unchanged, and this topic has been an active research area since the early days of neural network research \citep{hecht-nielsenALGEBRAICSTRUCTUREFEEDFORWARD1990}. For example in an MLP, permuting the neurons in one layer together with the outgoing weights preserves the function, as does similarly permuting the channels in a convolutional neural network. Non-linear activation functions induce further scaling symmetries \citep{godfreySymmetriesDeepLearning2022}; e.g. for a constant $\lambda \geq 0$, $\text{ReLU}(\lambda x) = \lambda \text{ReLU}(x)$, which means scaling the input weights to a neuron and applying the inverse scaling to the outgoing weights again preserves neural network's function. In addition to these static symmetries, other forms of symmetries might arise from the structure of the data, or dynamically during periods of training \citep{zhaoFindingSymmetryNeural2024}. 

Accounting for these symmetries in a weight-space learning task by using architectures with the correct inductive biases, through data augmentation \citep{shamsianImprovedGeneralizationWeight2024}, or by removing the symmetries by mapping neural networks to canonical representations \citep{pittorinoDeepNetworksToroids2022} as we will do can reduce the effective dimensionality of the problem and make weight-space learning more efficient. 

\subsubsection{Neural Networks as Graphs (Chapter \ref{section:wsl})}

A neural network can naturally be modeled as a graph, and this has fueled a recent line of work building graph neural networks (GNNs) that take as input other neural networks \citep{kofinasGraphNeuralNetworks2024,limGraphMetanetworksProcessing2023,kalogeropoulosScaleEquivariantGraph2024}. The nodes often correspond to the neurons in an MLP or channels in a convolutional network and the edges to the weights. Used with the appropriate positional encodings, this graph formalism provides an effective way of handling the permutation symmetries of various kinds of neural networks including transformers and neural networks with residual connections, and has also been extended to account for scaling symmetries \citep{kalogeropoulosScaleEquivariantGraph2024}. Furthermore, since a GNN is not restricted to graphs with a certain structure, weight-space GNNs have the additional benefit that the same GNN can be used to process different neural networks, even those with different architectures altogether.

As part of our flows, we process neural networks using GNNs, more specifically the Relational Transformer architecture \citep{diaoRelationalAttentionGeneralizing2023,kofinasGraphNeuralNetworks2024} that incorporates an attention mechanism with edge updates. 

\subsubsection{Generative Modeling with Flow Matching (Chapter \ref{section:flow_models})}

Flow matching (FM) \citep{lipmanFlowMatchingGenerative2023,albergoStochasticInterpolantsUnifying2023,liuFlowStraightFast2022,tongImprovingGeneralizingFlowbased2023} generalizes diffusion models with a more flexible framework and a simple simulation-free regression objective. Given two sample-able marginal distributions $p_0$ and $p_1$, a coupling $(x_0, x_1)$ is sampled from a joint distribution, and $x_t$ from the intermediate distribution $p_t(x_t \mid x_0, x_1)$ conditioned on the endpoints with $t$ sampled within the interval $[0,1]$. A neural network (velocity model) is then trained to predict the conditional velocity $u_t(x_t \mid x_0, x_1)$, such as $x_1 - x_0$. Marginalizing this linear vector field over the joint coupling then results in a more complex vector field that transforms $p_0$ to $p_1$, which is estimated by solving a differential equation with the velocity given by the trained velocity model. This flow matching framework can also be extended to data with more general geometries such as Riemannian manifolds \citep{chenRiemannianFlowMatching2023}. 

We will train our models using flow matching, with a Gaussian prior and the posterior samples collected through the neural networks' optimization trajectories, experimenting with different couplings. Our slightly modified flow matching training setup is described in more detail in Section \ref{sec:fm_training}.

\subsubsection{Flows with Different Geometries (Chapter \ref{chapter:method})}

We propose three flow models that handle the underlying geometric structure in different ways, which enables us to evaluate the effect of geometric considerations more precisely. We consider ReLU MLPs, and first align all neural networks to the same reference network using the re-basin operation \citep{ainsworthGitReBasinMerging2023,penaReBasinImplicitSinkhorn2023} that permutes a network's weights to minimize the loss barrier between two networks.  To handle the scaling symmetries, we build on the canonicalization procedure of \citep{pittorinoDeepNetworksToroids2022}, where each neuron's incoming weights are normalized and its outgoing weights are inversely scaled to preserve the function being computed; the last layer is further normalized globally in classification networks. This gives the neural network a product geometry, with the bias vectors as Euclidean vectors, intermediate neurons' incoming weight vectors on the hypersphere, and the last layer (optionally) on the hypersphere as a whole. 

We then compare our three flows: first a Euclidean flow that ignores the scaling symmetries and models each weight vector in Euclidean space, then a Normalized flow with the weights embedded in the product geometry but with the velocity field defined in Euclidean space (i.e. inside the hyperspheres), and finally a Geometric flow with the velocity field defined on the product geometry as well using the Riemannian Flow Matching framework \citep{chenRiemannianFlowMatching2023}. 

\subsubsection{Overview of Results (Chapter \ref{chapter:results})}

We evaluate our flows on a variety of tasks after training them on samples obtained with gradient-based optimization methods and show that for small networks on relatively easier tasks, they can directly generate weights matching or sometimes exceeding the performance of weights optimized with gradient-based methods. On more complex tasks and larger models, while direct generation does not match the quality of optimized weights, Bayesian model averaging over a number of samples leads to comparable predictive performance. Then we show that a flow trained on weights from one task can be transferred to another task, either by using the sampled weights as learned initializations, or by guiding the sampling process with task gradients. Our results finally demonstrate that in particular the Euclidean and Geometric flows can also generalize to different architectures for the same base task, and that there are still gains to be made by further scaling up our models. 

\section{Contributions}

Our main contributions are as follows:
\begin{itemize}
    \item We utilize flow matching in weight-space for the first time. Our Euclidean flow learns to transport weights in Euclidean space directly, then we model the neural networks on a product manifold by removing their scaling symmetries to build our Normalized flow, and finally in our Geometric flow we utilize the Riemannian Flow Matching framework to define a vector field over this product manifold. 
    \item We empirically validate our proposed methods by learning to approximate the posterior distributions of neural networks of various sizes on different tasks. We observe that they can generate samples competitive with weights optimized via gradient-based methods, and that can still be scaled up for better performance. 
    \item We show that our flows can also generalize to different architectures, such as by sampling more accurate weights for a larger mode. This indicates they do not learn to model just a single posterior distribution, but instead learn to model weights on the underlying task more generally. 
    \item Finally we apply guidance with task gradients during sampling as an instance of transfer learning and show that a flow built with weights trained on one dataset can be used to sample weights that are accurate on another dataset. 
\end{itemize}

% !TeX root = ../main.tex
% Add the above to each chapter to make compiling the PDF easier in some editors.

% \chapter{Background \& Related Work}\label{chapter:Background}

% !TeX root = ../main.tex
% Add the above to each chapter to make compiling the PDF easier in some editors.

\chapter{Bayesian Deep Learning}\label{section:bayesian_dl}

A primary use case for a generative model over neural network weights is in Bayesian deep learning, where it can allow efficient inference by transporting the prior distribution to the posterior. Thus, to motivate the rest of the discussion, we first give an overview of concepts from Bayesian deep learning. Section \ref{section:bayesian_concepts} is a general introduction, followed by a review of inference methods (Laplace approximations, variational inference, MCMC-based methods) typically used for Bayesian neural networks.

\section{General Concepts} \label{section:bayesian_concepts}

Bayesian deep learning aims to quantify the uncertainty in neural networks through probability distributions over their parameters, rather than obtaining a single solution by an SGD-like optimization method (refer to \citep{mackayBayesianMethodsAdaptive1992, nealBayesianLearningNeural1996} for foundational work and \citep{goanBayesianNeuralNetworks2020,arbelPrimerBayesianNeural2023} for more recent reviews). With the \textit{posterior} distribution $p(\theta \vert \calD)$ over weights $\theta$ conditioned on the dataset $\calD$, predictions are obtained via \textit{Bayesian model averaging}:
\begin{equation} \label{eq:model_averaging}
    p(y \vert x, \calD) 
    = \bbE_{\theta \sim p(\theta \vert \calD)} \left[ p(y\vert x, \theta) \right]
    = \int p(y \vert x, \theta) p(\theta \vert \calD) d\theta,
\end{equation}
where the prediction is an expectation over the posterior. Note that since the forward pass through the model $p(y \vert x, \theta)$, is deterministic, the uncertainty in predictions results solely from the uncertainty over parameters. To bring things together, Bayesian inference over neural network weights consists of three steps:
\begin{enumerate}
    \item Specify prior $p(\theta)$.
    \item Compute/sample posterior $p(\theta \vert \calD) \propto p(\calD \vert \theta) p(\theta)$.
    \item Average predictions over the posterior. 
\end{enumerate}
The last step only requires forward passes through the model and thus is straightforward. The first two steps require deeper consideration. 

\section{Priors}

Specifying a prior mainly consists of two choices: specifying an architecture, and specifying a probability distribution over the weights. A standard choice for a prior distribution is an isotropic Gaussian, which is an uninformative prior but therefore widely applicable and flexible. 

Different architectural decisions also induce different distributions over functions, even if the flattened weight vectors are identical, meaning that the choice of an architecture further specifies a prior in function space. As a simple example, keeping the depth and width of a neural network constant, even just changing the activation function from a ReLU to a sigmoid results in a different distribution of functions. The functional distribution can also be specified in a more deliberate way; e.g. a translation-invariant convolutional neural network, or a group-equivariant network \citep{cohenGroupEquivariantConvolutional2016} puts probability mass only on functions satisfying certain equivariance constraints depending on the task at hand. 

We keep this discussion short since the choice of a prior has tangential impact in the rest of the presentation, and we refer to recent reviews such as \citep{fortuinPriorsBayesianDeep2022} for a more detailed treatment of Bayesian neural network priors. 

\section{Inference} \label{section:bayesian_inference}

Inference in BNNs is a rich research area covering a wide range of methods. In this section we give an overview of key methods used to approximate BNN posteriors, and refer to works such as \citep{arbelPrimerBayesianNeural2023} and \citep{murphyProbabilisticMachineLearning2023} for more comprehensive treatments. 

\subsection{Laplace Approximation}

A Laplace approximation to the posterior \citep{mackayPracticalBayesianFramework1992} is essentially a Gaussian distribution centered at the MAP estimate of the posterior, with the covariance matrix given by the log likelihood's Hessian's inverse. This intuitively corresponds to a local second-order Taylor expansion around the MAP estimate. 

A Laplace approximation is simple and fast, but ignores the posterior's multi-modality as it only captures a single mode. Nevertheless, it has been an active topic of research \citep{daxbergerLaplaceReduxEffortless2021} due to its simplicity and for use cases benefiting from local posterior estimates. 

\subsection{Variational Inference}

In a variational approximation, the posterior is estimated with a parametric distribution whose parameters are learned to minimize the KL-divergence between the posterior and the variational approximation; e.g. learning the mean and covariance matrix of a Gaussian distribution. The parameters are often learned to optimize the evidence lower bound (ELBO) using reparametrization tricks \citep{kingmaAutoEncodingVariationalBayes2022b,blundellWeightUncertaintyNeural2015a}. 

Variational approximations trade off expressivity for tractable optimization and thus can learn more complex distributions than a Gaussian, e.g. by accounting for symmetries in a neural network's posterior landscape \citep{gelbergVariationalInferenceFailures2024}. Nevertheless, KL-based objectives may require a large number of likelihood evaluations and often suffer from mode collapse, where the approximation captures only one mode of the distribution \citep{felardosDesigningLossesDatafree2023}. 

\subsection{Sampling-Based Inference}

Rather than parametrize or approximate the posterior distribution, sampling-based methods which our flow is a part of, attempt to sample from the posterior directly. Two main approaches to sampling are Monte Carlo methods \citep{barbuMonteCarloMethods2020} such as Markov chain Monte Carlo or importance sampling, and map-based methods \citep{marzoukSamplingMeasureTransport2016} such as normalizing flows \citep{rezendeVariationalInferenceNormalizing2015}.

\subsubsection{Monte Carlo Methods}

Monte carlo methods are designed around a few core ideas such that the samples are asymptotically from the true posterior, and are often used to compute expectations of the form
\begin{equation} \label{eq:expectation}
    \bbE_{x \sim p(x)} \left[ h(x) \right].
\end{equation}
A simple approach is Importance Sampling (IS) \citep{kahnMethodsReducingSample1953}, where samples from a proposal distribution $q$ are re-weighted with $w(x) = p(x)/q(x)$ to estimate the expectation in Equation \ref{eq:expectation}:
\begin{equation}
    \bbE_{x \sim q(x)} \left[ \frac{p(x)}{q(x)} h(x) \right]
    = \int q(x) \frac{p(x)}{q(x)} h(x) dx = \int p(x)h(x) dx = \bbE_{x \sim p(x)} \left[ h(x) \right].
\end{equation}
While IS does not require sampling from the posterior, the rate of convergence for the estimate depends on the variance of the IS weights, i.e. how well the proposal distribution aligns with the true posterior. 

Alternatively, Markov chain Monte Carlo (MCMC) methods such Hamiltonian Monte Carlo \citep{nealMCMCUsingHamiltonian2011}, Stochastic Gradient Monte Carlo \citep{maCompleteRecipeStochastic2015}, or Langevin Monte Carlo \citep{robertsExponentialConvergenceLangevin1996} construct a Markov chain of samples that eventually converges (mixes) to the true posterior. This is often achieved through a Metropolis-Hastings correction step \citep{metropolisEquationStateCalculations1953,hastingsMonteCarloSampling1970} where a proposed step $z$ from the current state $x$ is accepted with probability 
\begin{equation}
    a(x,z) = \min \left\{ 1, \frac{p(z)}{p(x)} \frac{q(z,x)}{q(x,z)} \right\}
\end{equation}
where $q(x, z)$ gives the transition probability from $x$ to $z$. 

While chain-based methods are heavily used in practice, they suffer from various shortcomings such as generating correlated rather than independent samples, and requiring a large number of likelihood evaluations some of which is later rejected. Measuring the convergence of a chain is also challenging, lacking concrete rules. 

\subsubsection{Map-Based Sampling}

Map-based methods \citep{marzoukSamplingMeasureTransport2016} are instead use a map $T:\R^n \to \R^n$ that transforms samples from the prior distribution $q$ to samples from the posterior $p$, such that the push-forward of the prior gives the posterior distribution: $T_\sharp  q = p$. Such a map can generate independent, uncorrelated samples without any likelihood evaluations which is a desirable property. 

The theory of map-based sampling is closely related to optimal transport (OT) \citep{peyreComputationalOptimalTransport2020,santambrogioOptimalTransportApplied2015} with a history going back to the 18th century \citep{mongeMemoireTheorieDeblais1781}, where the goal is to find the map $T$ that minimizes some cost $c(x, T(x))$ such as Euclidean distance. Under the later relaxation of the problem by Kantorovich \citep{kantorovitchTranslocationMasses1958} to joint distributions rather than deterministic maps to minimize the cost over the joint distribution with marginal constraints, the optimal map is guaranteed to exists given mild conditions. 
 
Nevertheless, while a map-based approach has benefits over an MCMC method such as providing less ambigous convergence criteria by reducing the problem to optimization, the optimal map is often hard to find, and therefore approximate maps are used instead. Such maps can be learned as static \citep{rezendeVariationalInferenceNormalizing2015} or continuous normalizing flows \citep{chenNeuralOrdinaryDifferential2018a} which have seen risign interest recently with developments such as flow matching which we now turn to.

% !TeX root = ../main.tex
% Add the above to ea  ch chapter to make compiling the PDF easier in some editors.

\chapter{Flow Models}\label{section:flow_models}

\section{Overview}

We train our flow using \textit{flow matching} \citep{lipmanFlowMatchingGenerative2023,albergoStochasticInterpolantsUnifying2023,liuFlowStraightFast2022}, which generalizes diffusion models with a more flexible design space. In this section we first formulate the flow matching objective (Sec. \ref{section:flow_matching}), explain the design choices it enables (Sec. \ref{section:design_choices}), and describe in more detail how likelihoods (Sec. \ref{section:computing_likelihoods}) can be computed using a flow model. 

\section{Flow Matching} \label{section:flow_matching}

Flow matching, first proposed in \citep{lipmanFlowMatchingGenerative2023,albergoStochasticInterpolantsUnifying2023,liuFlowStraightFast2022} (see \citep{lipmanFlowMatchingGuide2024} for a recent comprehensive overview), aims to solve the problem of \textit{dynamic transport}, i.e. finding a time-dependent vector field to transport the source (prior) distribution $p_0$ to the target (data) distribution $p_1$. More formally, the vector field $u_t: [0,1] \times \R^d \to \R^d$ leads to the ordinary differential equation (ODE)
\begin{equation} \label{eq:ode}
    dx = u_t(x) dt
\end{equation}
and induces a \textit{flow} $\phi: [0,1] \times \R^d \to \R^d$ that gives the solution to the ODE at time $t$ with starting point $x_0$, such that 
\begin{align}
    \frac{d}{dt} \phi_t(x_0) &= u_t(\phi_t(x_0)) \\
    \phi_0(x_0) &= x_0.
\end{align}
Starting with $p_0$, transformed distributions $p_t$ can then be defined using this flow with the push-forward operation
\begin{equation}
    p_t := [\phi_t]_\# (p_0)
\end{equation} 
and the instantaneous change in the density satisfies the \textit{continuity equation}
\begin{equation}
    \frac{\partial p}{\partial t} = - \nabla \cdot (p_t u_t)
\end{equation}
which means that probability mass is conserved during the transformation. With these formulations, we say the vector field $u_t$ \textit{generates} the \textit{probability path} (also called \textit{interpolant}) $p_t$.

\subsection{The Objective}

The formulation above could also be applied to traditional continuous normalizing flows (CNFs) \citep{chenNeuralOrdinaryDifferential2018a}, and flow matching is an instantiation of CNFs. However, continuous normalizing flows have in the past been trained using objectives which required solving and then backpropagating through the ODE, such as KL-divergence or other likelihood-based objectives, which made training costly. This problem was later addressed with diffusion models and their simpler regression objectives such as score matching and denoising \citep{sohl-dicksteinDeepUnsupervisedLearning2015, songScoreBasedGenerativeModeling2021a,hoDenoisingDiffusionProbabilistic2020} that proved to be very effective. The flow matching objective is also formulated as a simulation-free regression objective, and is more flexible than the diffusion objectives. 

As explained in Section \ref{section:flow_matching}, the goal in flow matching is to learn a vector field $v_\theta: [0,1] \times \R^d \to \R^d$ parametrized by a neural network.\footnote{For conciseness, we interchangeably use the subscripts for vector fields to denote time ($u_t(x)$) and parameters ($v_\theta(t,x)$).} If we know the ground truth vector field $u$ and can sample from the intermediate $p_t$'s, we can directly optimize the flow matching objective 
\begin{equation} \label{eq:fm_objective}
    \Lfm := \bbE_{t \sim \U(0,1), x_t \sim p_t(x)} \Vert v_\theta(t, x_t) - u_t(x_t) \Vert ^2
\end{equation}
by first sampling a time point $t$ and then $x_t \sim p_t$. However in practice, we neither have a closed form expression for $u$ nor can sample from an arbitrary $p_t$ without integrating the flow. 

The \textit{conditional flow matching} (CFM) framework first introduced in \citep{lipmanFlowMatchingGenerative2023} and then extended in \citep{tongImprovingGeneralizingFlowbased2023} solves this problem by formulating the intermediate probability paths as mixtures of simpler paths, 
\begin{equation}
    p_t(x) = \int p_t(x \mid z) q(z) dz
\end{equation}
where $z$ is the conditioning variable and $q(z)$ a distribution over $z$ (e.g. with $z := (x_0, x_1), q(z) = p_0(x_0)p_1(x_1), u_t(x \mid z) = x_1 - x_0$, and $p_t(x \mid z) = \N(x \mid (1-t)x_0 + tx_1, \sigma^2)$). Then similar to how $p_t$'s were generated by the vector field $u_t$, the conditional probability paths $p_t(x \mid z)$ are generated by conditional vector fields $u_t(x \mid z)$, and as shown in \citep{tongImprovingGeneralizingFlowbased2023} $u_t$ can be decomposed in terms of these conditional vector fields as 
\begin{equation}
    u_t(x) = \bbE_{z \sim q(z)} \frac{u_t(x \mid z) p_t(x \mid z)}{p_t(x)}.
\end{equation}
Then similar to Equation \ref{eq:fm_objective}, we have the conditional flow matching objective 
\begin{equation} \label{eq:cfm_objective}
    \Lcfm := \bbE_{t \sim \U(0,1), z \sim q(z), x_t \sim p_t(x \mid z)}
    \Vert v_\theta(t, x_t) - u_t(x_t \mid z) \Vert ^2.
\end{equation}
That is, we first sample a conditioning variable $z$, and then regress to the \textit{conditional} vector field $u_t(x \mid z)$. Thus we obtain a tractable objective by defining sample-able conditional probability paths and a tractable conditional vector field. Moreover, as shown in \citep{tongImprovingGeneralizingFlowbased2023}, the FM and CFM objectives are equivalent up to a constant and therefore
\begin{equation}
    \nabla_\theta \Lfm = \nabla_\theta \Lcfm,
\end{equation}
meaning we do not lose the expressive power of the FM objective by regressing only to the conditional vector fields. As we will show in Section \ref{section:design_choices}, the choice of these conditional probability paths, vector fields, along with the conditioning variable itself, makes the flow matching approach particularly flexible.

\subsection{Training} \label{sec:fm_training}

To sum up the discussion in the previous section, a step of training a flow model using the CFM objective (Equation \ref{eq:cfm_objective}) proceeds as follows:
\begin{enumerate}
    \item Sample $t \sim \U(0,1), z \sim q(z)$, and $x_t \sim p_t(x \mid z)$.
    \item Compute $\Lcfm = \Vert v_\theta(t, x_t) - u_t(x_t \mid z) \Vert ^2$.
    \item Update $\theta$ with $\nabla_\theta \Lcfm$.
\end{enumerate}
An equivalent approach with the loss computed in the space our data lies is to predict the target $x_1$ rather than the velocity and compute the loss as $\Vert v_\theta(t,x_t) - x_1 \Vert^2$, and the velocity as $v_\theta(t, x_t) - x_0$ with $x_0$ the starting point for integration. 

\subsection{Couplings and Conditional Paths} \label{section:design_choices}

With this framework established, the three main design choices for building a flow matching model are choosing the coupling $q(z)$, the conditional ``ground truth'' vector field $u_t(x \mid z)$, and the conditional probability paths $p_t(x \mid z)$. Starting with an arbitrary source distribution $p_0$ and target distribution $p_1$, \citet{tongImprovingGeneralizingFlowbased2023} propose three different ways of constructing conditional paths from couplings between $p_0$ and $p_1$, of which we focus on two (independent and optimal transport couplings). In all setups, the condition variable $z$ corresponds to a pair $(x_0, x_1)$ of source and target points. 

\textbf{Independent Coupling.} The simplest way of obtaining is to sample independently from $p_0$ and $p_1$; i.e. $q(z) = p_0(x_0) p_1(x_1)$, with the conditional paths and the vector field defined as 
\begin{align} 
    p_t(x \mid z) &= \N(x \mid (1-t)x_0 + t x_1, \sigma^2) \label{eq:cond_path} \\
    u_t(x \mid z) &= x_1 - x_0. \label{eq:cond_vector_field}
\end{align}
The conditional paths and the coupling defined this was are easily easy to sample from, but have undesirable properties such as crossing paths which which can lead to high variance in the ground truth vector field for a specific point and time. Moreover in practice, independent couplings can lead to curved paths that incur higher integration errors, as there is no notion of straightness considered in this formulation.

\textbf{Optimal Transport.} To obtain straighter and shorter paths that are easier to integrate, \citet{tongImprovingGeneralizingFlowbased2023} propose to use the static 2-Wasserstein optimal transport map $\pi$ as the coupling; i.e.
\begin{equation}
    q(z) = \pi(x_0, x_1),
\end{equation}
with the conditional paths and vector field defined as in Equations \ref{eq:cond_path} and \ref{eq:cond_vector_field}. The flow model thus obtained solves the dynamic optimal OT problem as $\sigma^2 \to 0$ (Proposition 3.4 in \citep{tongImprovingGeneralizingFlowbased2023}). However, computing the exact OT map for the entire dataset is challenging, especially in high dimensions as in our problem. It can instead be approximated using mini-batches \citep{fatrasMinibatchOptimalTransport2021}. This means at the end the OT problem is solved only to an approximation, but nevertheless results in straighter paths that cross less often, since intuitively an $x_0 \sim p_0$ is more likely to be coupled with $x_1 \sim p_1$ closer to it rather than an $x_1$ chosen uniformly random.

\section{Riemannian Flow Matching} \label{sec:riemannian_fm}

When our data lies on a manifold, flow matching can be extended to define a flow over the manifold as well. Such an approach can both lead to a better scalable generative model by reducing the effective dimensionality of the problem, and make learning easier by introducing a strong inductive bias to the problem.  Discrete and continuous normalizing flows have previously been adapted to Riemannian manifolds \citep{gemiciNormalizingFlowsRiemannian2016,mathieuRiemannianContinuousNormalizing2020,louNeuralManifoldOrdinary2020}, and in this section we begin with a brief overview of Riemannian manifolds (we refer to textbooks on the topic such as \citep{johnm.leeIntroductionRiemannianManifolds2018} for a more rigorous treatment), and then explain how an ODE over a Riemannian manifold can be learned following the framework of Riemannian flow matching \citep{chenRiemannianFlowMatching2023}.\footnote{The presentation in this section is additionally based on the Geometric Generative Models tutorial by Joey Bose, Alexander Tong, and Heli Ben-Hamu at the 2024 Learning On Graphs Conference.} 

\subsection{A Brief Review of Riemannian Manifolds}

\begin{figure}[t!]
    \includegraphics[width=\linewidth]{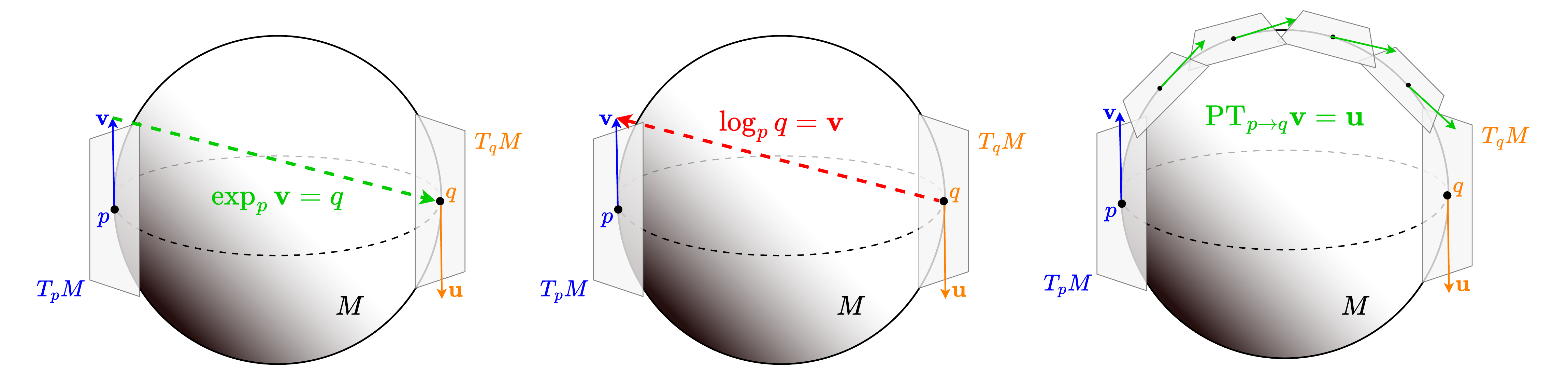}
    \caption{\label{fig:manifolds}\textbf{Visualizing the exponential, logarithmic, and parallel transport maps on a manifold.} The unique geodesic curve $\gamma$ between $p$ and $q$ with $\dot \gamma(0) = \mathbf{v}$ corresponds to the great circle of the sphere coinciding with the upper boundary in the diagrams.}    
\end{figure}

A smooth \textit{manifold} $M$ is a smooth topological space that is locally Euclidean. The local Euclidean structure is represented by \textit{charts} mapping open sets $U \subset M$ to $\R^n$. A set of charts covering the entire manifold is called an \textit{atlas}, and smoothness arises from the transitions between overlapping charts being smooth functions. Each point $x \in M$ is equipped with a tangent space $T_x M$ that is a vector space containing vectors tangent to $M$, such as velocities. In addition to being a smooth manifold, a Riemannian manifold also contains a Riemannian metric $g$ that defines an inner product $\langle u, v \rangle_g = u^T G v$ for $u, v \in T_x M$ and positive definite matrix $G$. A metric most importantly enables us to define angles and distances over the manifold as $\Vert u \Vert_g = \sqrt{\langle u, u\rangle_g} = \sqrt{u^T G u}$ and $\cos \theta = \frac{\langle u, v \rangle_g}{\Vert u \Vert_g \Vert v \Vert_g}$. 

A \textit{curve} is a smooth function $\gamma: [0, 1] \to M$, and tangent vectors $v \in T_x M$ can be expressed as time derivatives $\dot \gamma(t)$ of curves with $\gamma(t) = x$. Using the metric $g$, we can measure the length of a curve by computing the length of the tangent vector at each point along the curve:
\begin{equation}
    \vert \gamma \vert =
    \int_0^1 \Vert \dot \gamma (t) \Vert_g^2 dt.
\end{equation}
The ``shortest'' curve in this sense connecting two points $x,y\in M$ is called a \textit{geodesic}. We can thus define a distance between two points as the length of a (not necessarily unique) geodesic connecting them. 

A manifold is also equipped with three operations that we will make use of: the \textit{exponential map}, the \textit{logarithmic} map, and \textit{parallel transport} (see Figure \ref{fig:manifolds} for a visual depiction). The exponential map $\exp_x: T_x M \to M$ at a point $x \in M$ maps a tangent vector $v$ to the point $\gamma(1) \in M$ where $\gamma$ is the unique geodesic satisfying $\gamma(0) = x$ and $\dot \gamma(0) = v$. The logarithmic map $\log_x: M \to T_x M$ maps a point $y \in M$ back to the tangent vector $v := \dot \gamma (0)$. It is generally the inverse of the exponential map. Finally the parallel transport map $\text{PT}_{x \to y}: T_x M \to T_y M$ transports a tangent vector $v := \dot \gamma(0)$ along the geodesic with $\gamma(1) = y$ keeping the lengths and angles between the transported vectors constant. Together, these three operations allow us to move along geodesics with tangent vectors as the velocities, and transport tangent vectors to the same tangent space to be able to compute distances between them, which are all steps required to build a flow model over a Riemannian manifold. 

\subsection{Flow Models over Manifolds}

Riemannian flow matching (RFM) \citep{chenRiemannianFlowMatching2023} extends the flow matching framework to Riemannian manifolds. As the three key design choices, we need to define a coupling $q$, conditional probability paths $p_t$, and a conditional vector field $u_t$. 

A coupling can be defined simply by sampling $x_0 \sim p_0$ and $x_1 \sim p_1$ independently, or with a mini-batch optimal transport map with the distances defined above using geodesics. For $t\in [0,1]$, the geodesic interpolation $x_t \sim p_t(x_t \mid x_0, x_1)$ (analogous to linear interpolation in Euclidean space) between $x_0$ and $x_1$ is computed as 
\begin{equation}
    x_t := \exp_{x_0}(t \log_{x_0} x_1)
\end{equation}
and the conditional vector field $u_t$ as
\begin{equation}
    u_t(x_t \mid x_0, x_1) := \frac{\log_{x_t}x_1}{1 - t}.
\end{equation}
Finally to solve the ODE, we compute one Euler integration step as
\begin{align}
    dx_t &= u_t(x_t)dt \\
    x_{t + 1} &= \exp_{x_t}(v_\theta(t, x_t) \Delta t)
\end{align}
where $\Delta t$ is the step size.

\section{Computing Likelihoods} \label{section:computing_likelihoods}

In earlier normalizing flows that aim to learn a static mapping between the two distributions \citep{rezendeVariationalInferenceNormalizing2015}, given source samples $x_0 \sim p_0$, likelihoods of the generated samples $z = f(x_0) \approx p_1$  can be computed exactly via the change of variables formula
\begin{equation} \label{eq:static_cov}
    \log p_1(z) = \log p_0(z) - \log \det \left\vert J_f(z) \right\vert
\end{equation}
where $J_f$ is the Jacobian of $f$. Thus, we can obtain exact likelihoods for the generated samples by taking the determinant of the Jacobian of the normalizing flow. Since Jacobian computations can be costly, this has motivated work on designing normalizing flows with easier to compute Jacobians, such as RealNVP \citep{dinhDensityEstimationUsing2017}. 

% Note that the terminology/notation etc are already set here
In a \textit{continuous normalizing flow} on the other hand, the \textit{instantaneous change of variables} formula \citep{chenNeuralOrdinaryDifferential2018a} defines the change in probability mass through time. Given that the vector field $v_t$ is continuous in $t$ and uniformly Lipschitz continuous in $\R^d$, it holds that
\begin{align} \label{eq:continuous_cov}
    \frac{d \log p_t(\phi_t(x))}{dt} &= - \divergence(v_t(\phi_t(x))) \\
                                     &= - \trace \left( \frac{d v_t(\phi_t(x))}{dt} \right)
\end{align}
where $\frac{d v_t(\phi_t(x))}{dt} =: J_v(\phi_t(x))$ is the Jacobian of the vector field. We integrate over time to compute the full change in probability:
\begin{equation} \label{eq:full_continuous_cov}
    \log p_1(\phi_1(x)) = \log p_0(\phi_0(x)) - \int_0^1 \trace(J_v(\phi_t(x))) dt.
\end{equation}
Then we can integrate the Jacobian trace of the vector field through time (simultaneously with sampling) to obtain exact likelihoods for the generated samples. 

\subsection{Faster Likelihoods Through Trace Estimation} \label{section:trace_estimation}

However, materializing the full Jacobian of the vector field can be prohibitively expensive, especially if the task is very high dimensional (as in our case) since the log determinant computation has a time complexity of $O(d^3)$ \citep{grathwohlFFJORDFreeformContinuous2018} without any restrictions on the structure of the Jacobian. 

To alleviate this problem, \citep{grathwohlFFJORDFreeformContinuous2018} propose to use the \textit{Hutchinson trace estimator} \citep{hutchinsonStochasticEstimatorTrace1990} for an unbiased estimate of the Jacobian trace of a square matrix: 
\begin{equation} \label{eq:hutchinson}
    \trace(J_v) = \bbE_{p(\epsilon)} \left[ \epsilon^T J_v \epsilon \right]
\end{equation}
where $p(\epsilon)$ is chosen such that $\bbE[\epsilon] = 0$ and $\Cov(\epsilon) = I$, typically a Gaussian or a Rademacher distribution. Then, we can use this estimator in place of the explicit trace computation in Equation \ref{eq:full_continuous_cov} and compute the likelihoods as
\begin{equation}
    \log p_1(\phi_1(x)) = \log p_0(\phi_0(x)) - \int_0^1 \bbE_{p(\epsilon)} \left[ \epsilon^T J_v(\phi_t(x)) \epsilon \right] dt.
\end{equation}

The performance benefit of using the Hutchinson trace estimator results from the fact that the Jacobian-vector product $J_v \epsilon$ can be computed very efficiently by automatic differentiation \citep{baydinAutomaticDifferentiationMachine2018}, giving the whole approach a time complexity of $O(d)$ only. Due to this significant performance improvement and being an unbiased estimate, the Hutchinson trace estimator has been widely used in the diffusion/flow model literature \citep{lipmanFlowMatchingGenerative2023,songScoreBasedGenerativeModeling2021a}.

% !TeX root = ../main.tex
% Add the above to each chapter to make compiling the PDF easier in some editors.

\chapter{Symmetries of Neural Network Weights}\label{section:geometry_of_nns}

Symmetries in data, such as rotation symmetries in molecules or translation symmetries in images, can be used to obtain useful inductive biases for ML models \citep{bronsteinGeometricDeepLearning2021,weilerEquivariantCoordinateIndependent2023}. By restricting the search space to functions that respect those symmetries, such inductive biases make training more data-efficient \citep{brehmerDoesEquivarianceMatter2024}. Since our data modality is neural network weights, taking the symmetries of neural networks into account can likewise make learning in weight space more effective. 

Given a neural network $f$ with parameters $\theta$, we define a \textit{symmetry} as an operation $\phi$ that leaves the function computed by the neural network unchanged; i.e. $f_{\phi(\theta)}(x) = f_\theta(x)$. We are further interested only in the static symmetries that hold for any $\theta$ and thus can more reliably be used as inductive biases, rather than dynamic symmetries specific to a certain value of $\theta$. Such symmetries of neural network weights have been studied for a long time \citep{hecht-nielsenALGEBRAICSTRUCTUREFEEDFORWARD1990}, and are still key considerations for understanding the loss landscape and training dynamics of neural networks \citep{breaWeightspaceSymmetryDeep2019a,simsekGeometryLossLandscape2021,limEmpiricalImpactNeural2024,zhaoIMPROVINGCONVERGENCEGENERALIZATION2024}.

\section{Permutation and Scaling Symmetries of Neural Networks} \label{sec:perm_and_scale_sym}

\begin{figure}[t!]
    \centering
    \includegraphics[width=\textwidth]{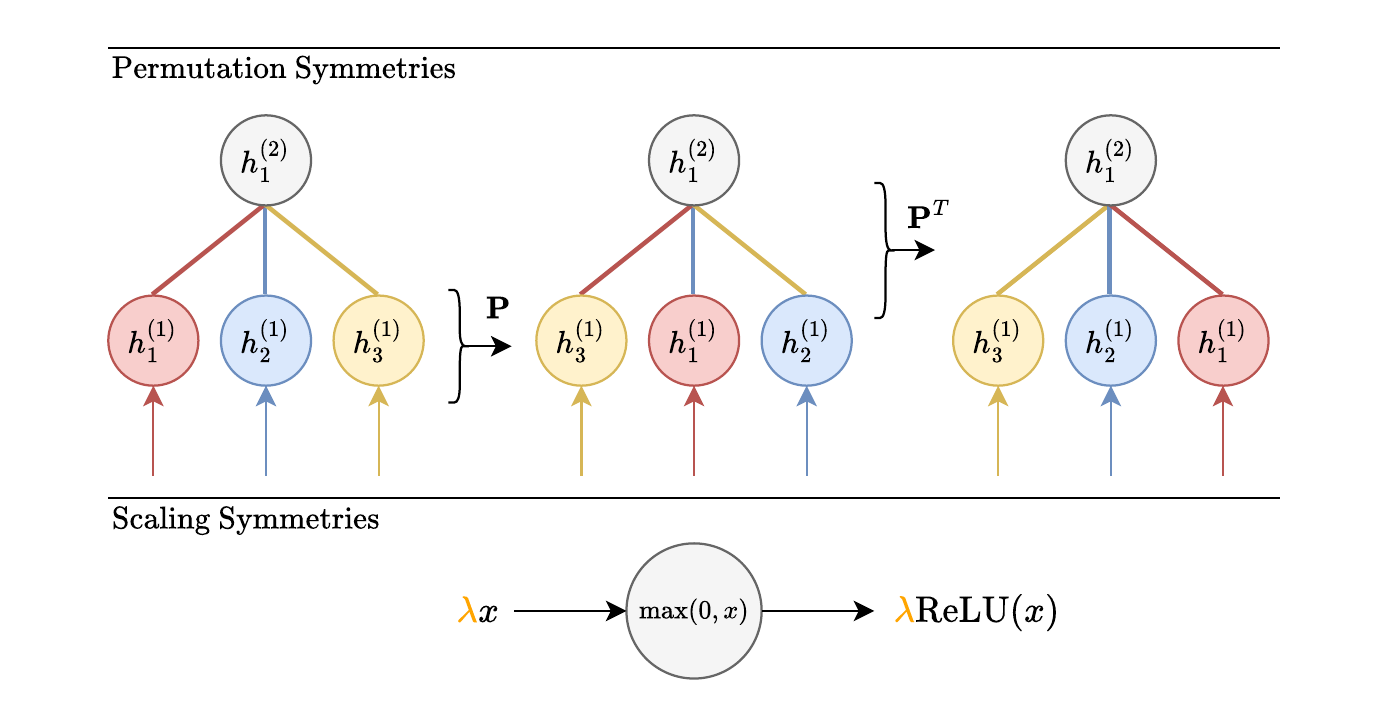}
    \caption{\label{fig:nn_sym}\textbf{Visual illustration of example permutation and scaling symmetries of neural networks.} \textit{Top}: Permuting the neurons at one layer and then applying the same permutation to the outgoing weights preserves the function being computed. \textit{Bottom:} For a ReLU activation, multiplying the input with a non-negative constant and the output with its inverse preserves the function. }    
\end{figure}

A typical neural network with non-linear activations has two main kinds of symmetries: \textit{permutation symmetries} and \textit{scaling symmetries}. Permutation symmetries arise from the connectivity structure of the neural network, and scaling symmetries mainly arise from the particular non-linearities. 

More formally, consider a two-layer MLP with weight matrices $\W_1, \W_2$ and element-wise activations $\sigma$; i.e. $f_\theta(x) = \W_2 \sigma (\W_1x)$, ignoring the biases for simplicity but the following discussion applies to the biases as well. Let $\bP$ be an arbitrary permutation matrix. We can permute the hidden neurons, and apply the same permutation to their outgoing weights as well to keep the function unchanged:
\begin{equation}
    f_\theta(x) = \W_2 \sigma (\W_1x) = \W_2 \bP^T \sigma (\bP \W_1x).
\end{equation}
Since $\sigma$ is applied element-wise, we have $\W_2 \bP^T  \bP \sigma (\W_1x)$ which preserves the function as $\bP^T  \bP = \mathbf{I}$. Similarly, convolutional neural networks' channels can be permuted while preserving the function. More generally, \citet{limGraphMetanetworksProcessing2023} show that the permutation symmetries of any neural network correspond to graph automorphisms of its neural DAG, constructed with each edge corresponding to a parameter (e.g. directly the computational graph for MLPs, or with each filter corresponding to a node for CNNs).

In addition to these permutation symmetries, element-wise activation functions such as ReLU introduce further scaling symmetries to neural networks. For example, for the ReLU activation it holds for any real number $\lambda > 0$ that $\text{ReLU}(\lambda x) = \lambda \text{ReLU}(x)$. This means the input weights to a layer with ReLU activations can be multiplied with a positive number, and the function will remain unchanged as long as the output weights are scaled down with the same factor. Although we will mostly consider ReLU networks in the following sections, it is worth noting that such scaling symmetries exist for activation functions besides ReLU as well. \citep{godfreySymmetriesDeepLearning2022} have shown that symmetries resulting from activation functions can be associated with different \textit{intertwiner groups}, and provide concrete examples of these groups for various activation functions. 

\section{Linear Mode Connectivity}

\begin{figure}[t!]
    \centering
    \includegraphics[width=0.6\textwidth]{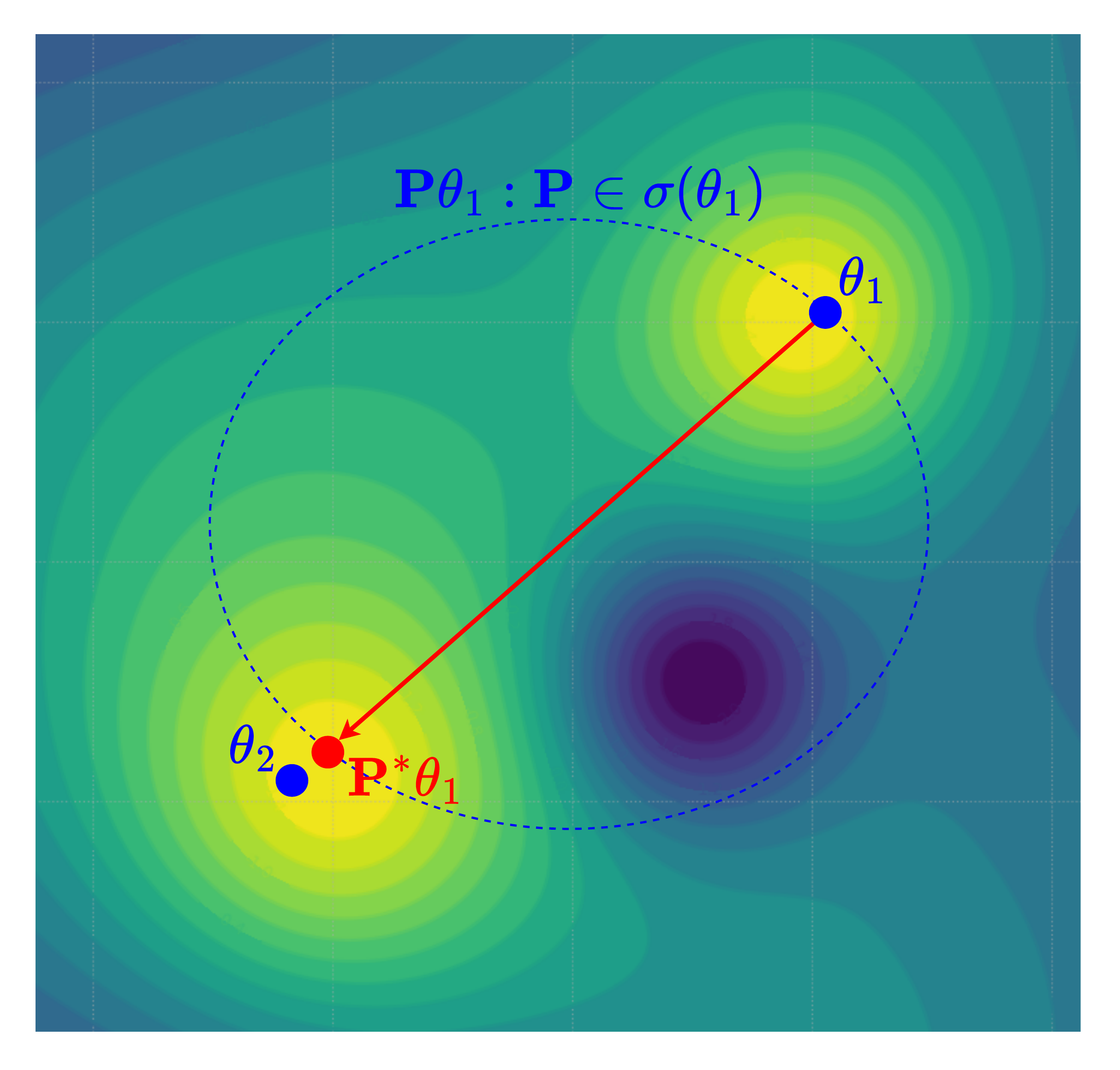}
    \caption{\label{fig:mode_conn}\textbf{Linear mode connectivity}. The hypothesis asserts that up to permutations, low-loss points in a neural network's loss landscape are linearly connected.}    
\end{figure}

Closely related to the literature on symmetries of neural network weights is the topic of \textit{linear mode connectivity}, concerned with finding (linear) low-loss paths between SGD-optimized weights. Finding such paths is useful for many downstream applications, such as ensembling neural networks \citep{garipovLossSurfacesMode2018a} by finding accurate weights with different representations without training, or model merging \citep{stoicaZipItMergingModels2024}.

\citet{garipovLossSurfacesMode2018a} first formulated the problem as finding parametrized curves between NN weights, minimizing the loss across the curve. It was later conjectured by \citet{entezariRolePermutationInvariance2022} that up to permutation symmetries, such low-loss paths are linear. The linear mode connectivity hypothesis has since then given rise to a fruitful research area \citep{ferbachProvingLinearMode2024,rossiPermutationSymmetriesBayesian2023,zhaoUnderstandingModeConnectivity2023}. 

For our purposes, modes being linearly connected would imply that finding the optimal permutations could effectively reduce the number of modes in the weight-space posterior, making it easier to approximate. Accounting for this multi-modality has been shown to improve the effectiveness of Bayesian neural networks \citep{sommerConnectingDotsModeConnectedness2024}. With this motivation, we next focus on the literature around finding such permutations. 

\section{Aligning Neural Networks}

The problem of finding a permutation of one neural network weights to obtain a linear low-loss path with another neural network, we call \textit{aligning} neural networks, has given rise to a high number of methods over years, the entirety of which could be a thesis in itself. For instance, \citep{ainsworthGitReBasinMerging2023} propose various approaches. The first is a data-based approach that matches the activations of models $A$ and $B$ of each layer,
\begin{equation}
    \bP^* = \arg\min_{\bP \in S_d} \sum_{i=1}^n \Vert \mathbf{Z}_A - \bP \mathbf{Z}_B \Vert ^2
\end{equation}
with $d$ dimensional activations $\mathbf{Z}$ for $n$ data points. This is an instance of the \textit{linear assignment problem} which can be solved in polynomial time \citep{crouseImplementing2DRectangular2016}. An alternative is to align the weights of the neural networks directly, which can again be reduced to a linear assignment problem and is more efficient since it requires no forward passes over the model, but sacrifices accuracy by ignoring the data. 

\citet{penaReBasinImplicitSinkhorn2023} propose a more flexible framework, making it possible optimize for any differentiable objective, and we also use their approach in the rest of our work. \citet{penaReBasinImplicitSinkhorn2023} start by relaxing the constraint of binary permutation matrix $\bP \in \Pi$ to obtain unconstrained $\mathbf{X} \in \R^{m\times n}$. Such a matrix can then be mapped to the space of binary permutation matrices via the \textit{Sinkhorn operator}:
\begin{equation} \label{eq:sinkhorn}
    S_\tau(\mathbf{X}) = \arg\max_{\bP \in \Pi} \langle \bP, \mathbf{X} \rangle_F + \tau h(\bP)
\end{equation}
where $F$ denotes the Frobenius norm and entropy $h(\bP) = - \sum \bP \log \bP$. Then the optimization is performed over $\R^{m \times n}$, and a binary permutation matrix is obtained through the Sinkhorn operator. 

The main advantage of the approach of \citet{penaReBasinImplicitSinkhorn2023} is that it can be used with arbitrary differentiable objectives. To align the parameters $\theta_A$ and $\theta_B$, with $\pi_\bP$ denoting the permutation applied to the weights, \citet{penaReBasinImplicitSinkhorn2023} propose three objectives. First is a straightforward weight-matching similar to \citep{ainsworthGitReBasinMerging2023}
\begin{equation}
    \mathcal{L}_{L2} := \Vert \theta_A - \pi_\bP (\theta_B) \Vert^2,
\end{equation}
followed by two data-based objectives. Minimizing the mid-point loss between the two weights as
\begin{equation}
    \mathcal{L}_{\text{Mid}} := \mathcal{L} \left( \frac{\theta_A + \pi_\bP (\theta_B)}{2} \right) 
\end{equation}
or the loss at a random intermediate point 
\begin{equation}
    \mathcal{L}_{\text{Rnd}} := \mathcal{L} \left( (1-\lambda)\theta_A + \lambda \pi_\bP (\theta_B) \right) 
\end{equation}
with $\lambda \sim \mathcal{U}(0,1)$. At the end, particularly the data-based losses result in more effective permutations than the method of \citep{ainsworthGitReBasinMerging2023}, and we choose to use the approach of \citet{penaReBasinImplicitSinkhorn2023} in the rest of our work also considering its flexibility. 

\section{Canonical Representations of Neural Networks}

This discussion around permutation and scaling symmetries of neural networks culminates with \textit{canonical representations} of neural networks, i.e. unique representations for each set of permutation/scale-symmetric neural networks, limiting our discussion to ReLU networks for simplicity. Following the work of \citep{pittorinoDeepNetworksToroids2022}, this can achieved in two steps given a set of neural networks $\{ \theta_i\}_i^N$:
\begin{enumerate}
    \item Align all neural networks to a single \textit{reference} neural network $\theta'$, using the approach of \citep{penaReBasinImplicitSinkhorn2023}. 
    \item For each intermediate layer $l$ and neuron $k$, scale down the incoming weights and biases by the norm of the incoming weight vector, $\vert w_k^l \vert^{-1}$, and the outgoing weights by $\vert w_k^l \vert$. Additionally for classification tasks, normalize the last layer's weights to $\sqrt{C}$, with $C$ the number of classes. This does not change the predicted label due to the argmax operation at the last layer. 
\end{enumerate}
With these two operations, the permutation and scaling symmetries are ``broken,'' as all the neural networks that compute the same function up to permutation and scaling are now mapped to the same point in weight space. Nevertheless, while the scaling symmetry is broken exactly, the permutation symmetry is broken up to an approximation since the alignment methods \citep{ainsworthGitReBasinMerging2023,penaReBasinImplicitSinkhorn2023} only output approximate solutions. For this reason, as we will describe in the next section, we use graph neural networks to fully account for the permutation symmetries. This canonicalization also gives a specific geometric structure to the set of neural network weights. Each neuron, characterized by its incoming weights, now lies on the unit hypersphere, and each layer in turn has a product geometry of hyperspheres. This enables the computation of geodesic paths and distances between neural networks. 

% !TeX root = ../main.tex
% Add the above to each chapter to make compiling the PDF easier in some editors.

\chapter{Weight-Space Learning}\label{section:wsl}

\begin{figure}[h!]
    \centering
    \includegraphics[width=0.8\linewidth]{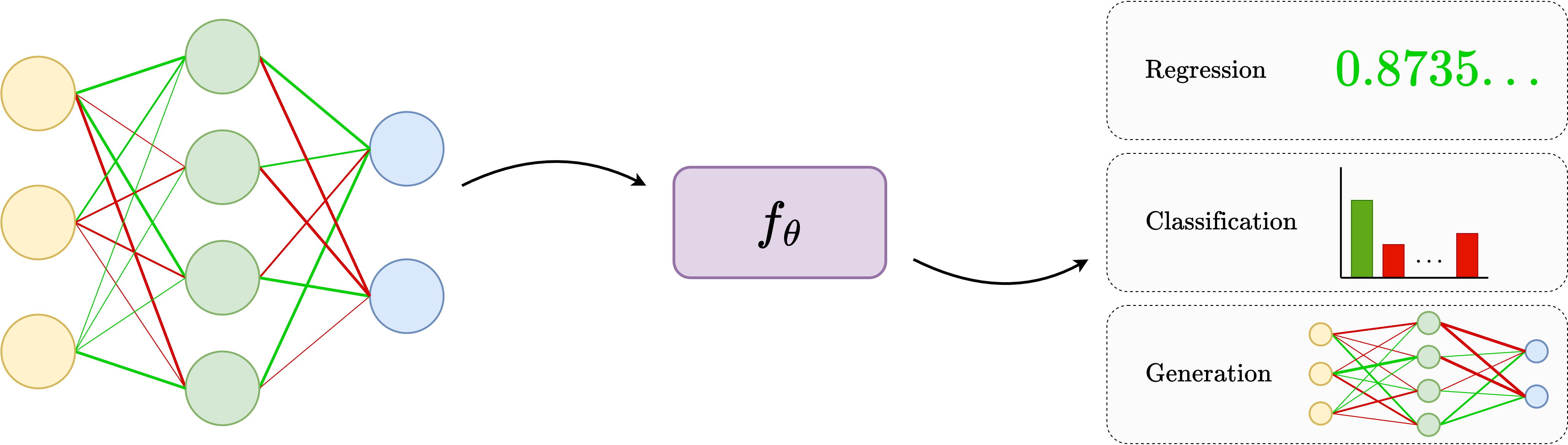}
    \caption{\label{fig:wsl} \textbf{Weight-space learning}. Neural network weights are processed using other neural neural networks for tasks such as regression (e.g. predicting the loss of unseen weights), classification (e.g. ), or generation (e.g. learned optimization).}    
\end{figure}

Training neural networks on other neural networks' weights has been an active research area for a long time \citep{haHyperNetworks2016, kruegerBayesianHypernetworks2018}, but has seen increasing interest recently with new neural network architectures being proposed \citep{limGraphMetanetworksProcessing2023,kofinasGraphNeuralNetworks2024,
zhouNeuralFunctionalTransformers2023,zhouUniversalNeuralFunctionals2024}. This wave of interest, combined with developments around other problems such as generative modeling, has led to a wider range of applications of weight-space learning, including generative modeling of neural network weights \citep{peeblesLearningLearnGenerative2022, erkocHyperDiffusionGeneratingImplicit2023} and machine unlearning using weight-space models \citep{rangelLearningForgetUsing2024}. In this section, we provide a review of recent work on weight-space architectures and weight-space generative models. 

\section{Architectures}

\subsection{Graph Neural Networks}

A graph neural network (GNN) \citep{wuGraphNeuralNetworks2022} takes as input a graph $(V,E)$ with nodes $n_i \in V$ and edges $e_{ij} \in E$ with $i,j$ node indices, and operate by iteratively updating the node and edge features of $d_V$ and $d_E$ dimensions respectively. Although edge features may be omitted, a general node and edge GNN update step can be expressed as
\begin{align} \label{eq:mpnn}
    n_i^{l + 1} &=
    \phi_N \left( n_i, \bigoplus_{j \in N_i} \
                    \phi_M(e_{ij}^l, n_i^l, n_j^l) \right) \\
    e_{ij}^{l+1} &= \phi_E \left(
        e_{ij}^l, e_{ij}^l, n_i^{l+1} , n_j^{l+1} 
        \label{eq:edge_updates}
    \right)
\end{align}
where $\phi_N, \phi_M, \phi_E$ are the node, message, and update neural networks, and $\bigoplus$ is a permutation-invariant aggregation operation. $N_i$ is the \textit{neighborhood} of node $i$ which the update message is aggregated over. While directly using the connectivity structure of the input graph is a typical choice, methods using the entire graph \citep{diaoRelationalAttentionGeneralizing2023} or dynamically learning a structure (known as graph rewiring) \citep{gutteridgeDRewDynamicallyRewired2023} also exist.   Note that the weights of the neural networks are shared across nodes/edges and update steps, making the number of parameters in a GNN independent of the input graph's size. 

Different GNNs are characterized by how they construct the neural networks $\phi_N, \phi_M, \phi_E$. Graph convolutional networks \citep{kipfSemiSupervisedClassificationGraph2016} set $\phi_M(e_{ij}^l, n_i^l, n_j^l) = \phi_M(n_j^l)$ and $\bigoplus$ to be averaging. Graph attention networks \citep{velickovicGraphAttentionNetworks2018a} replace the average with self-attention. Transformers \citep{vaswaniAttentionAllYou2017a} can also be classified as GNNs, where $N_i = V$ and $\phi_M$ is self-attention.

\subsection{Graph Neural Networks in Weight-Space}

Since a neural network can be represented as a graph via its computational graph, GNNs make for a natural choice in constructing weight-space architectures, and there has been a recent line of work in this direction \citep{zhouNeuralFunctionalTransformers2023, limGraphMetanetworksProcessing2023, kofinasGraphNeuralNetworks2024,kalogeropoulosScaleEquivariantGraph2024}. We build on the architecture of \citep{kofinasGraphNeuralNetworks2024} and explain its workings in this section. 

\subsubsection{Construcing Graphs from Neural Networks}

An MLP with $L$ layers, consisting of weight matrices $\{ \W^{(1)},...,\W^{(L)} \}$ and bias vectors $\{ \bb^{(1)},...,\bb^{(L)}\}$ with $\W^{(l)} \in \R^{d_l \times d_{l-1}}$ and $b^{l}\in \R^{d_l}$ can be considered a graph directly following its computational graph, with nodes corresponding to the neurons and edges to the parameters. Then in its simplest form, edge features are individual parameters, and node features are the individual bias components, with $d_V = d_E = 1$, although including additional features such as ``probe features'' that correspond to the activations of certain inputs are also possible. This construction, since it follows the computational graph, respects the permutation symmetries of nodes in subsequent layers as described in Chapter \ref{section:geometry_of_nns}.

For convolutional neural networks (CNNs), the permutation symmetries occur at the level of individual channels. Permuting the filters in one layer permutes the channels at the subsequent layer, and permuting the filters of the subsequent layer with the same permutation preserves the function being computed. We can obtain this symmetry as a graph by associating nodes with individual channels and edges with the filters, where a single edge's features correspond to the flattened weights in a zero-padded filter to make all edge features the same size. 

\subsubsection{Learning over Neural Network Graphs} \label{sec:learning_graphs}

A standard GNN can be used to learn a function over NN weights with the graph constructed as above. However, not all GNNs incorporate edge features. \citet{kofinasGraphNeuralNetworks2024} propose two architectures, one extending the \textit{Principal Neighborhood Aggregation} network (PNA) \citep{corsoPrincipalNeighbourhoodAggregation2020} with edge updates, and another a slightly updated version of a \textit{Relational Transformer} \citep{diaoRelationalAttentionGeneralizing2023}. 

\textbf{PNA} \citep{corsoPrincipalNeighbourhoodAggregation2020} builds a message-passing scheme by combining various aggregators and scalers to obtain the aggregation operation
\begin{equation}
    \bigoplus = 
    \left[ \begin{array}{c}
        I \\ 
        S(D, \alpha = 1) \\ 
        S(D, \alpha = -1)
        \end{array} \right]
    \otimes
    \left[ \begin{array}{c}
        \texttt{mean} \\
        \texttt{std} \\
        \texttt{max} \\
        \texttt{min}
        \end{array} \right]
\end{equation}
where 
\begin{equation}
    S(d, \alpha) = \left( \frac{\log(d+1)}{\delta} \right)^\alpha
\end{equation}
scales the messages to reduce the effect of exponential changes in the messages. Updates then follow typical message-passing as in Equation \ref{eq:mpnn}, including edge features. The original formulation does not include edge updates, but \citet{kofinasGraphNeuralNetworks2024} add an update mechanism as in Equation \ref{eq:edge_updates}, as well as feature-wise linear modulation \citep{brockschmidtGNNFiLMGraphNeural2020} based on the edge features. 

The \textbf{Relational Transformer} \citep{diaoRelationalAttentionGeneralizing2023} is an extension of the traditional Transformer \citep{vaswaniAttentionAllYou2017a} to graphs with edge features. Node and edge updates follow the typical message-passing operations in Equations \ref{eq:mpnn}, \ref{eq:edge_updates}. Unlike typical attention where QKV vectors are obtained through node features alone, \citet{diaoRelationalAttentionGeneralizing2023} construct the QKV vectors by concatenating node and edge features; i.e.
\begin{equation}
    q_{ij} = [n_i, e_{ij}]\W^Q \quad
    k_{ij} = [n_i, e_{ij}]\W^K \quad
    v_{ij} = [n_i, e_{ij}]\W^V 
\end{equation}
where each weight matrix has two components, one for edges and one for nodes to obtain
\begin{equation}
    q_{ij} = \left( n_i \W_N^Q + e_{ij}\W_E^Q \right) \quad
    k_{ij} = \left( n_i \W_N^K + e_{ij}\W_E^K \right) \quad
    v_{ij} = \left( n_i \W_N^V + e_{ij}\W_E^V \right).
\end{equation}
The graph can be taken to be fully-connected, ignoring the original structure, but the architecture can be adapted to other forms of connectivity just by changing which nodes and edges each update is conditioned on. 

\section{Weight-Space Generative Models}

One of the earliest approaches to using a neural network to learn a generative model in weight-space is the \textit{Bayesian Hypernetworks} of \citet{kruegerBayesianHypernetworks2018}. A Bayesian hypernetwork converts noise to a sample weight $\theta$ from the approximate posterior $q(\theta)$. Making the hypernetwork invertible makes it possible to compute the log-determinant of its inverse Jacobian, similar to a normalizing flow. The hypernetwork and the base network can then be trained using backpropagation as a single model, where the hypernetwork outputs weights and the primary network is used to evaluate the ``true'' likelihood in a differentiable way. 

Although learning a generative model only with access to a likelihood function is still a very active research area (e.g. works such as \citep{tongImprovingGeneralizingFlowbased2023}), simulation-free regression objectives such as score/flow-matching have seen increasing interest, as they are cheaper to optimize and can avoid certain failure modes of likelihood-based objectives such as mode-seeking. This interest has also led to various applications of such methods in weight-space.

Using the denoising diffusion objective, \citet{peeblesLearningLearnGenerative2022} train a GPT-2 model \citep{radfordLanguageModelsAre2019} (omitting causal masking) over neural network weights, where each token is a vector of weights. The model, named \textbf{G.pt}, is trained using checkpoints from a large number of training runs augmented with permutations to obtain different representations of the same function. The model can then also be ``prompted'' for specific losses and succeeds at generating weights that correlate with the prompted losses. 

In an alternative approach, \citet{schurholtScalableVersatileWeight2024} aim to train an autoencoder that learns to map neural network weights to a lower-dimensional latent space, which can then facilitate various downstream tasks including generation. Each complete weight vector is tokenized and split into chunks, and each chunk has a separate latent representation. New weights are then sampled by fitting a kernel density estimator around the embeddings of the weights given as a prompt. The samples can then be iteratively refined by using the best samples as the new prompt and repeating this procedure.

% !TeX root = ../main.tex
% Add the above to each chapter to make compiling the PDF easier in some editors.

\chapter{Building Flows in Weight-Space} \label{chapter:method}

We now bring the preceding background together and describe how we can build flows in weight-space. We propose three alternative formulations based on how the scaling symmetries are handled: first a Euclidean flow defined in weight-space directly, and then the Normalized and Geometric flows with the weights normalized to remove the scaling symmetries, differing on whether the vector field is also defined in Euclidean space (Normalized flow) or on the manifold of normalized weights (Geometric flow). We conclude by describing by how we train and sample from our flows. 

\section{Training With and Without Samples}

Although simulation-free methods to train generative models with score/flow-matching objectives only given access to an unnormalized density function exist \citep{akhound-sadeghIteratedDenoisingEnergy2024}, we train our models on samples obtained using typical gradient-based optimization of neural networks. This is an easier setup and allows us to measure the impact of various design choices more clearly, and potentially also benefit from publicly available datasets of neural network weights \citep{schurholtModelZoosDataset2022,peeblesLearningLearnGenerative2022}. Thus to train our flow models, we independently train a number of neural networks on the base task and intermittently save the weights along each optimization trajectory, discarding the initial steps based on validation loss. 

\section{Flows with Different Geometries}

\begin{figure}[h!]
    \centering
    \includegraphics[width=\textwidth]{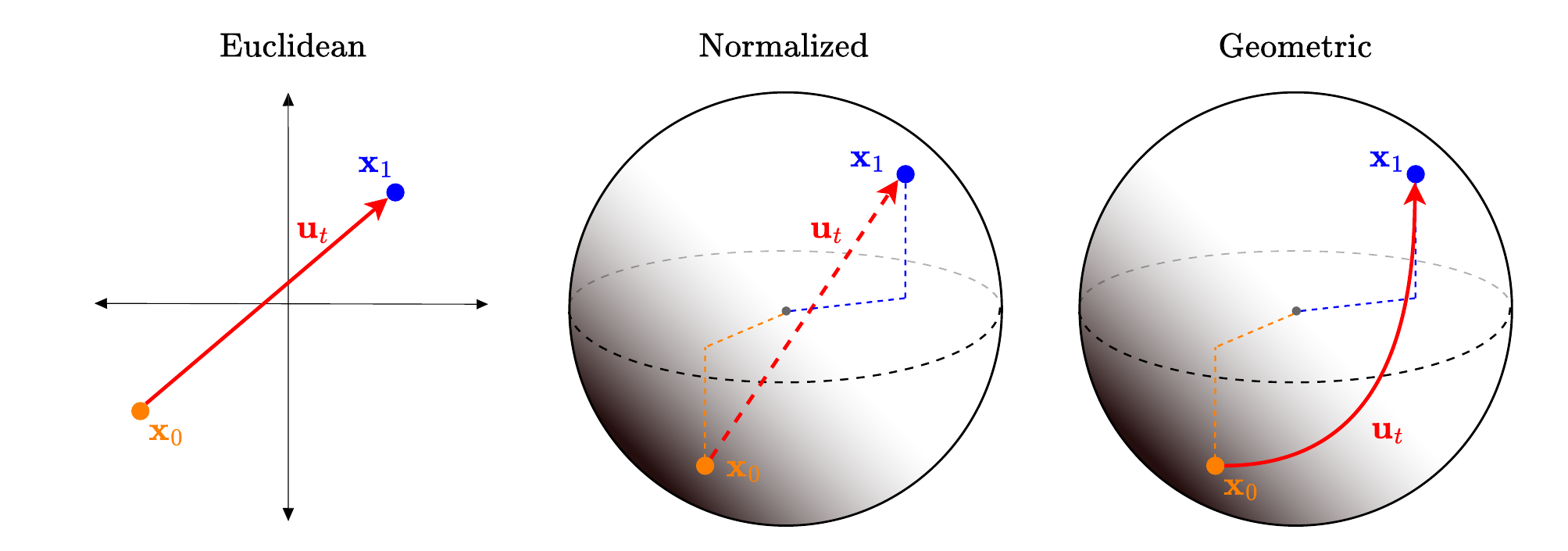}
    \caption{\label{fig:flow_types}\textbf{Flows with different geometric structures.} \textit{Left:} Euclidean flow defined over in weight-space without modifying the weights. \textit{Center:} Normalized flow after removing scaling symmetries, with the vector field defined as in Euclidean space. \textit{Right:} Geometric flow, with the vector field defined over the particular geometry as well.}
\end{figure}

The first step in designing a flow is to determine the space our data lies in. We propose three different approaches with different geometric properties, illustrated in Figure \ref{fig:flow_types}: first a Euclidean flow operating directly over the trained weights, and then two different methods operating with the normalized weights as described in Section \ref{section:geometry_of_nns} following the process of \citep{pittorinoDeepNetworksToroids2022}. The last two methods further differ on whether the vector space is also over these normalized weights or not. 

\subsection{Euclidean Flow}

Our first method we call the Euclidean flow is defined in Euclidean space, and the trained weights are used directly, without any transformations except permutation alignment (re-basin) to a common reference. Following the presentation in Section \ref{section:flow_matching}, we define the ground truth vector field as $x_1 - x_0$. Such a flow can be straightforwardly applied to any neural network, but ignores the scaling symmetries resulting from activations such as ReLU. 

\begin{figure}[t!]
    \centering
    \includegraphics[width=0.9\textwidth]{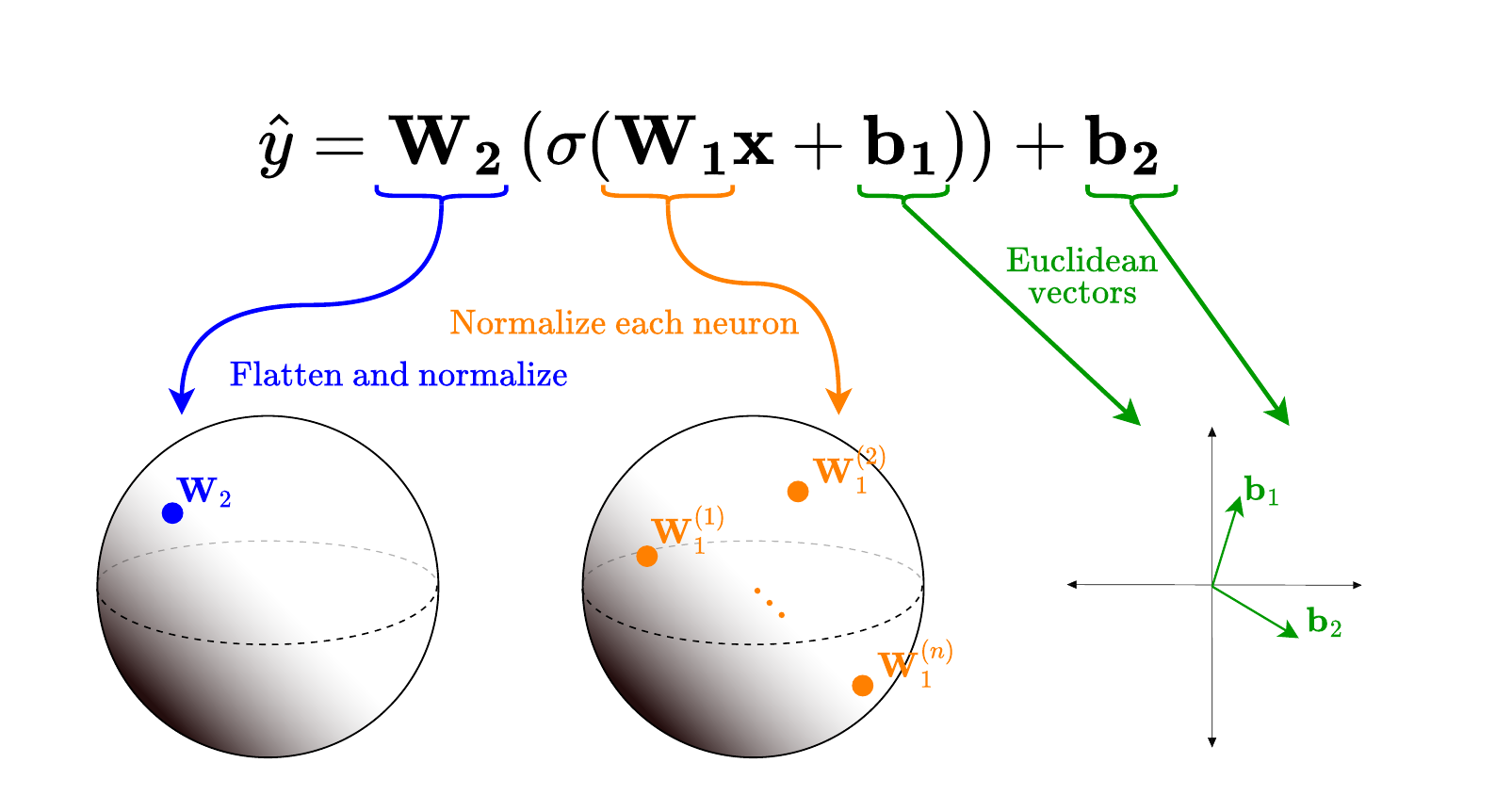}
    \caption{\label{fig:canonicalization}\textbf{Removing scaling symmetries from ReLU networks.} Each intermediate neuron lies on a unit hypersphere, as well as the last layer as a whole. Bias vectors are also rescaled to preserve the neural network's function but remain Euclidean vectors.}
\end{figure}

\subsection{Normalized Flow}

A simple modification to the Euclidean flow above is to first apply the normalization procedure of \citep{pittorinoDeepNetworksToroids2022} as shown in Figure \ref{fig:canonicalization}. For ReLU networks, we normalize the incoming weight vector and bias of each intermediate neuron, and scale the outgoing weight vector up by the same normalization factor to preserve the function being computed by the neural network. Since there is no such symmetry for the last layer, we normalize the entire last layer weights for a classification network as that would preserve the output class. 

This procedure embeds neural network weights in a product geometry. Each intermediate neuron's incoming weight vector, and the last layer in its entirety, now lie on unit hyperspheres of different dimensions. The bias vectors have no such structure and we still treat them as Euclidean vectors. Our Normalized flow works with source and target distributions on this product geometry, but the vector field is still defined in Euclidean space; i.e. inside the hyperspheres for neurons and in Euclidean space for biases. This is an intermediate step between the previous Euclidean flow and the Geometric flow we will describe next.  

\subsection{Geometric Flow}

Since hyperspheres, as well as the Euclidean space and a product of Riemannian manifolds, are Riemannian manifolds, we can model vector fields over them using the framework of Riemannian flow matching \citep{chenRiemannianFlowMatching2023} (Section \ref{sec:riemannian_fm}). For the bias vectors in Euclidean space, we again define $u_t := x_1 - x_0$ and $x_t = t x_1 + (1-t) x_0$. For intermediate neurons and the last layer on different hyperspheres, we have $x_t = \exp_{x_0}(t \log_{x_0}x_1)$ and $u_t = \log_{x_t}(x_1) / (1-t)$. This formulation, while computationally more expensive, has the added potential benefit that all inputs including the intermediate points $x_t$ to our model will lie on this particular geometry, reducing the effective dimensionality of the problem. 

While we focus only on ReLU networks, this geometric formulation can be extended to other non-linearities as they also induce different kinds of scaling symmetries \citep{godfreySymmetriesDeepLearning2022}. 

\section{Model Architecture}

For each of our three flows, we experimented with the PNA and Relational Transformer architectures and decided to use the Relational Transformer model due to its superior performance with fewer parameters. We thus model the vector field using a Relational Transformer with edge updates \citep{diaoRelationalAttentionGeneralizing2023,kofinasGraphNeuralNetworks2024}, and embed the time by appending it to each node and edge feature. For an MLP as the base model, each edge in the graph corresponds to a single weight, and nodes to individual bias values. We project each node and edge feature to $d_E$ dimensions using an MLP, append $t$ to each node and edge, and run a number of node/edge update steps before projecting the final node/edge states back to one dimension using an MLP. To distinguish nodes at different layers, we add layer-specific learned positional embeddings to edge features. 

\section{Training}

We train our models using the conditional flow matching \textbf{objective} (Equation \ref{eq:cfm_objective}), but rather than predict the velocity $u_t$, we predict the target point $x_1$ and compute the velocity during integration using the initial point $x_0$ and intermediate point $x_t$. This makes training easier to analyze as the error in weight-space is more directly informative, and we can constrain the outputs to the particular geometry for the Normalized and Geometric flows. 

\begin{wrapfigure}{r}{0.4\textwidth}
    \centering
        \includegraphics[width=\linewidth]{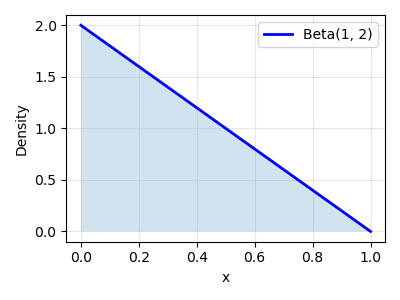}
    \caption{\label{fig:beta}\textbf{PDF of $\text{Beta}(1,2)$.} We sample $t\sim \text{Beta}(1,2)$ rather than uniformly when training our flow model.}
\end{wrapfigure}

As \textbf{prior} distributions $p_0$, we use isotropic Gaussian distributions of different variances, ensuring that the prior distribution is the same as the distribution used to initialize the weights while training the base model, although this is not a strict requirement and our flow models can be trained with arbitrary priors. We also experiment with independent and mini-batch optimal transport \textbf{couplings}, although especially in high-dimensions the bias induced by the mini-batch approximation to optimal transport can be significant \citep{fatrasMinibatchOptimalTransport2021}. 

As another modification over the standard flow matching framework, we sample \textbf{time} $t \in (0,1)$  during training not uniformly, but following $\text{Beta}(1,2)$ (Figure \ref{fig:beta}). Since $x_t$ with smaller $t$ are further away from $x_1$, estimation error increases for smaller time values. Sampling $t \sim \text{Beta}(1,2)$ results in a larger number of parameter updates for smaller $t$ relative to larger $t$, which is a more effective distribution of the computational budget since the predictions for larger $t$ are closer to $x_1$ and thus converge in fewer steps. Importantly, sampling $t \sim \text{Beta}(1,2)$ does not bias the training process, since a prediction at a certain time is independent of the model behavior at other times. 

\section{Sampling}

To sample from our flow models, we solve the ODE (Equation \ref{eq:ode}) using an Euler solver with step sizes depending on the complexity of the task, although higher-order solvers can also be used without any modification to the rest of the setup. We also experiment with \textbf{stochastic sampling} by adding noise before computing Euler updates, on the entire trajectory or a subset of it \citep{karrasElucidatingDesignSpace2022}. 

\subsection{Guidance}

We can also guide the sampling process with loss gradients from the base task \citep{wangProteinConformationGeneration2024,kulyteImprovingAntibodyDesign2024,yuForceGuidedBridgeMatching2024a}, so that for base model $f$ and velocity model $v_\theta$ Euler updates take the form 
\begin{equation}
    x_{t + \Delta t} = x_t + \left( v_\theta(t, x_t) + \lambda \nabla_{x_t}\mathcal{L}(f, x_t) \right) \Delta t
\end{equation}
where $\mathcal{L}(f, x_t)$ is computed with training data points from the base task (note that $x_t$ are neural network weights from the base task), and $\lambda$ is a coefficient controlling the strength of this guidance. If the task loss $\mathcal{L}$ is computed with mini-batches, this guidance also adds an implicit stochasticity to each step of the sampling process.

% !TeX root = ../main.tex
% Add the above to each chapter to make compiling the PDF easier in some editors.

\chapter{Experimental Results}\label{chapter:results}

We now evaluate our three flows and various choices on different models and datasets. The main experimental outcomes can be summarized as follows:
\begin{itemize}
    \item As a sanity check, we can train a flow between two Gaussian distributions, and stochastic sampling helps capture the variance in the posterior (Section \ref{sec:gaussian_flow}).
    \item On an easier task with a small model, all three flows can directly sample weights matching and even exceeding the quality of optimized weights (Section \ref{sec:uci_classification}).
    \item The Geometric and Normalized flows' sample qualities plateau after a smaller number of integration steps, indicating they learn shorter paths than the Euclidean flow (Section \ref{sec:uci_classification}).
    \item On a more difficult task and larger models, while individual samples perform worse than optimized weights, Bayesian model averaging with a small number of samples results in comparable performance (Section \ref{sec:sample_quality_diversity}).
    \item A similar sample quality can also be transferred to other tasks through guiding with task gradients during sampling (Section \ref{sec:task_generalization}).
    \item Using the samples weights as initialization to train models on a different task leads to faster convergence (Section \ref{sec:task_generalization}).
    \item The Euclidean and Geometric flows successfully generalize to a larger architecture on the same dataset, indicating they learn properties not only of the specific architecture but the underlying task (Section \ref{sec:arch_generalization}).
    \item There is still expected performance gains from using larger models than the ones we have used in our evaluations (Section \ref{sec:scaling}).
\end{itemize}

We describe the detailed experimental setups in Appendix \ref{appendix:experimental_setups}.

\section{Euclidean Flow Between Two Gaussians} \label{sec:gaussian_flow}

\begin{figure}[h!]
    \centering
    \begin{subfigure}{0.47\linewidth}
        \centering
        \includegraphics[width=\linewidth]{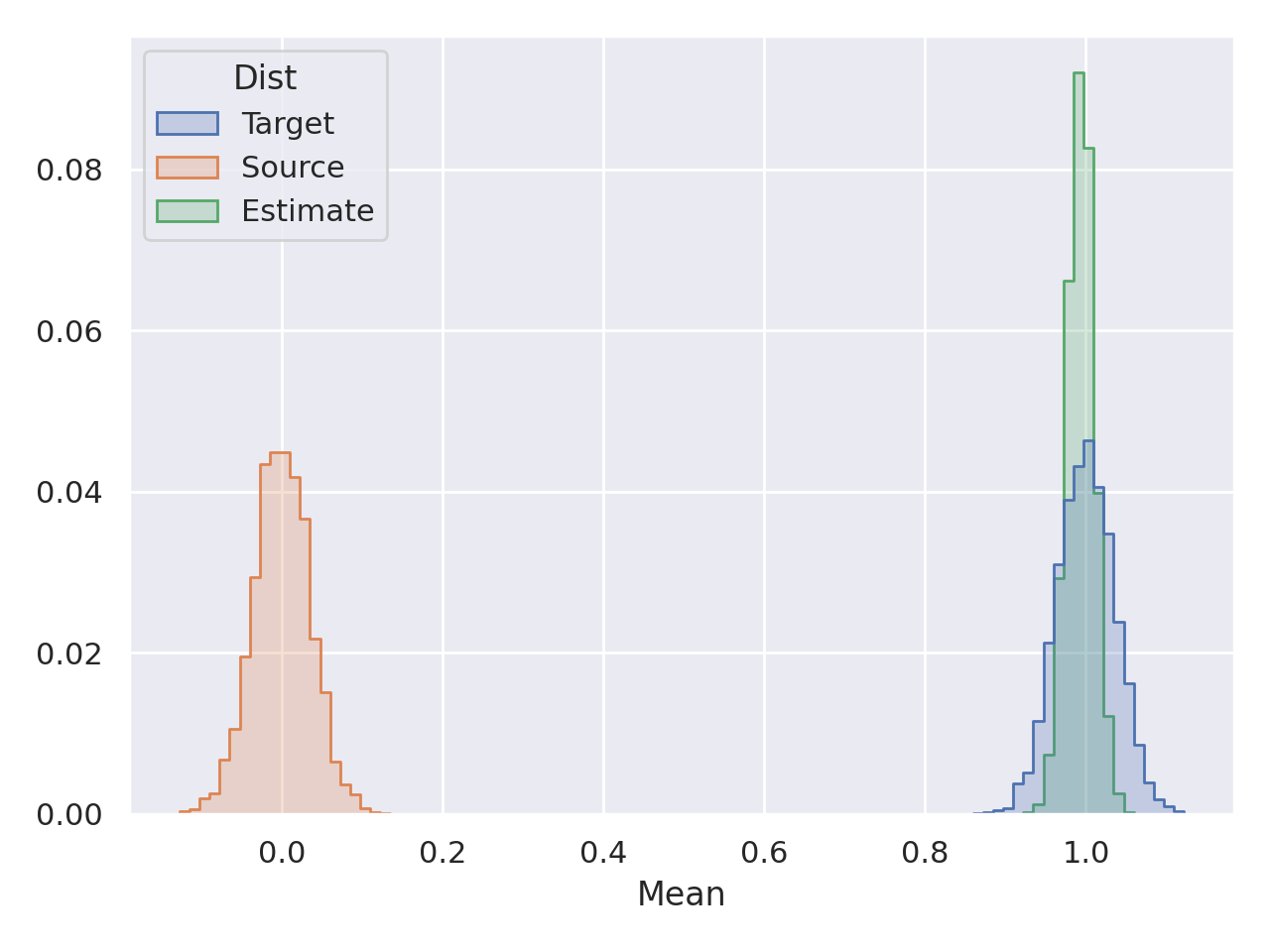}
        \caption{Deterministic sampling $(\varepsilon = 0)$}
        \label{fig:gaussian_deterministic}
    \end{subfigure}
    \begin{subfigure}{0.47\linewidth}
        \centering
        \includegraphics[width=\linewidth]{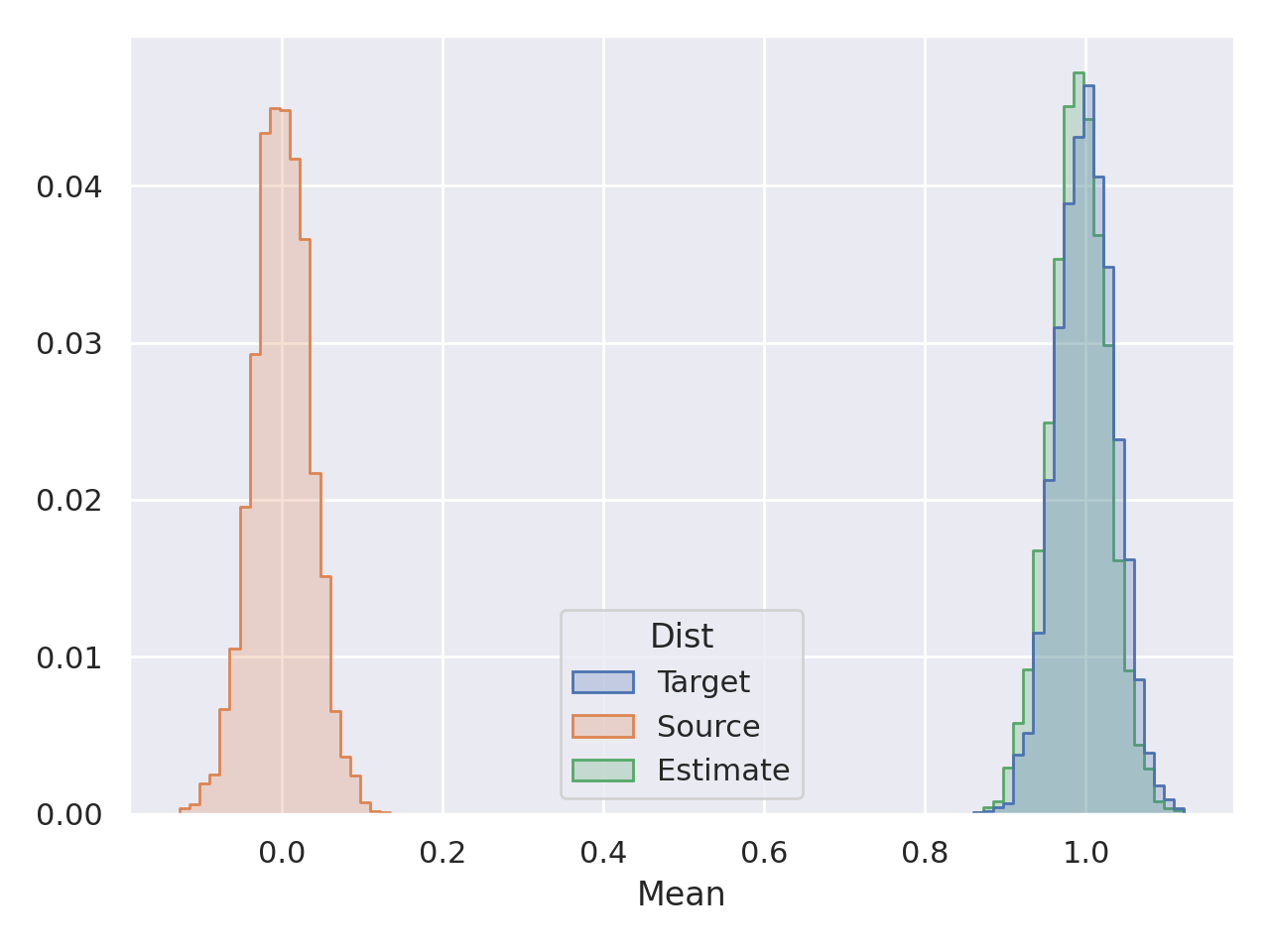}
        \caption{Stochastic sampling $(\varepsilon = 0.05)$}
        \label{fig:gaussian_stochastic}
    \end{subfigure}
    \caption{\label{fig:gaussian-results}\textbf{Histograms for the means of the weights generated by a Euclidean flow trained between two Gaussian distributions.} The flow fails to capture the variance in the target distribution with deterministic sampling, but this is corrected by stochastic sampling with $\varepsilon = 0.05$.} 
\end{figure}

To verify our approach of learning a flow model in weight-space, we begin our evaluation the toy task of learning a flow between two Gaussian distributions. The neural network is a small MLP with 30 input, two output dimensions, and two hidden layers of 16 neurons. We sample $X_0 \sim p_0 := \N(0, \mathbf{I})$ and $X_1 \sim p_1 := \N(1, \mathbf{I})$, and train our Euclidean flow to map $p_0$ to $p_1$ with independent coupling $q(x_0, x_1) = p_0(x_0)p_1(x_1)$. This is a relatively simpler task than learning over actual weights since each dimension of the weight vectors is sampled independently. 

Figure \ref{fig:gaussian-results} shows histograms of the means of the weights sampled from the flow with 100 Euler steps, and either deterministic or stochastic $(\varepsilon=0.05)$ sampling. Independent of the sampling method used, the flow covers the high-density center of the target distribution well, but the weights sampled deterministically fail to capture the variance in the target distribution. Stochastic sampling however appears to correct for this over-saturation and leads to more diverse samples. Overall, these results validates the feasibility of learning a flow model in weight-space using graph neural networks, and we move on to tasks involving actual learned weights. 

\begin{figure}[t!]
    \centering
    \begin{subfigure}{0.47\linewidth}
        \centering
        \includegraphics[width=\linewidth]{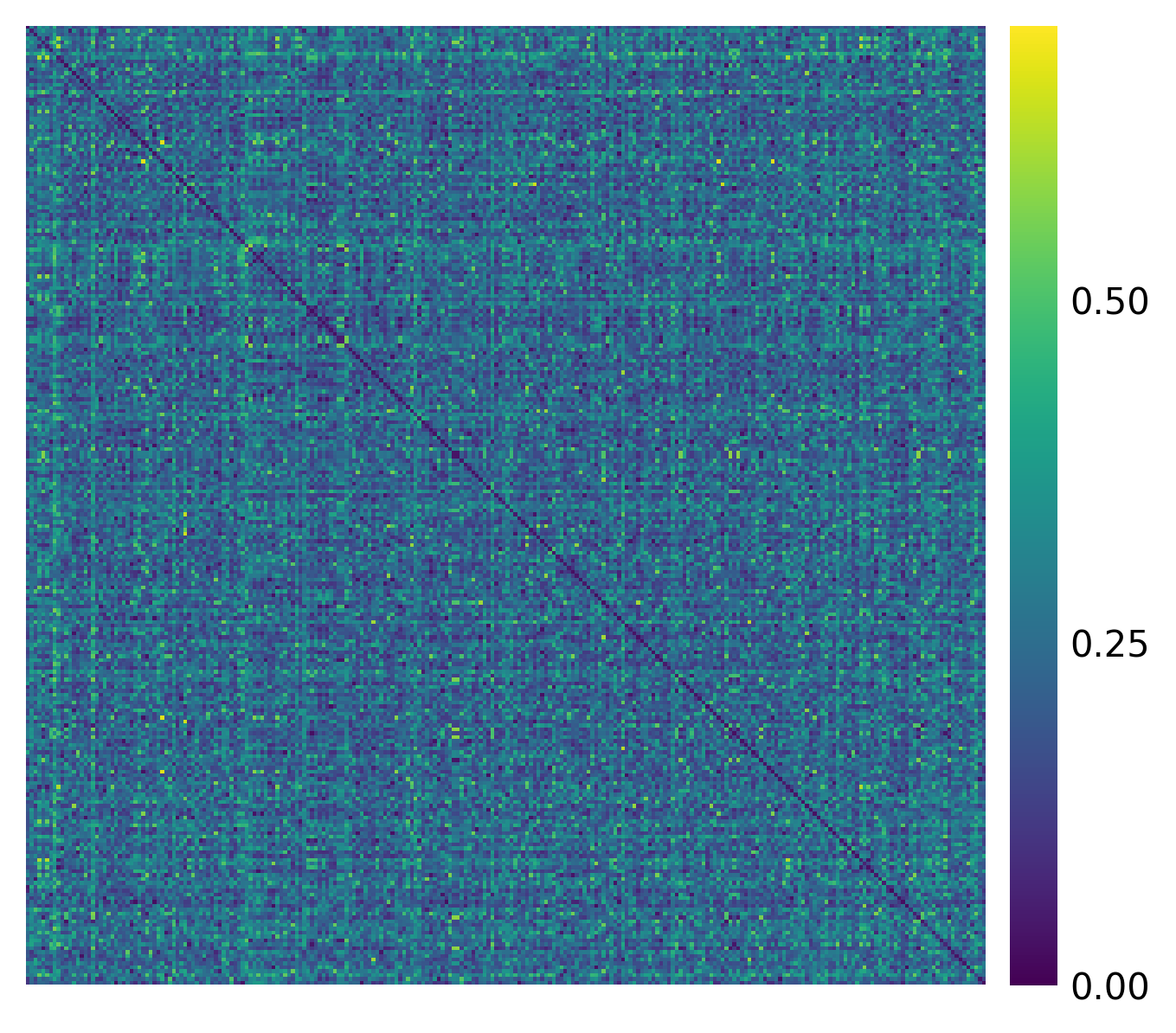}
        \caption{Unaligned (0.244 $\pm$ 0.009)}
        \label{fig:uci_unaligned}
    \end{subfigure}
    \begin{subfigure}{0.47\linewidth}
        \centering
        \includegraphics[width=\linewidth]{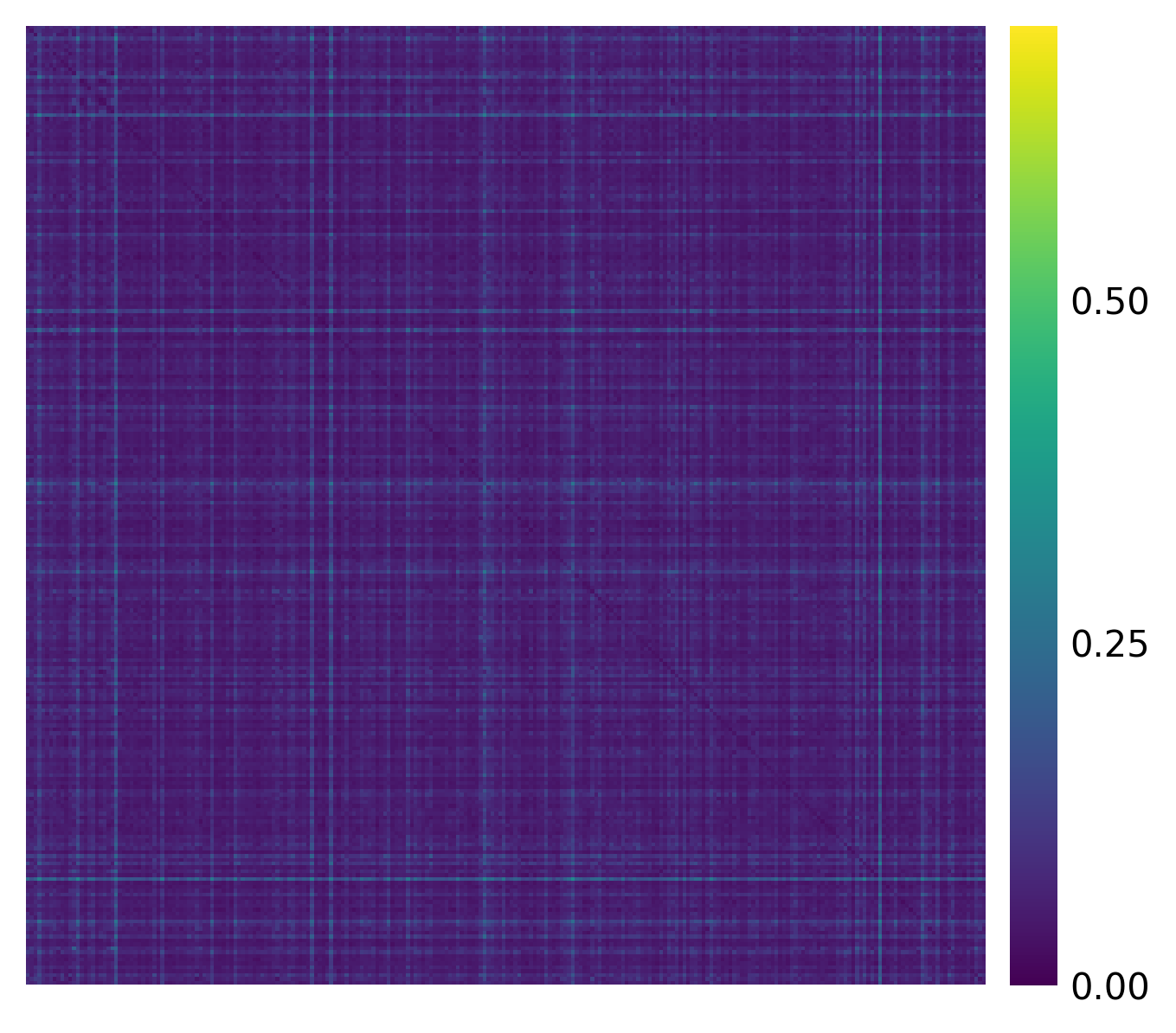}
        \caption{Aligned (0.070 $\pm$ 0.001)}
        \label{fig:uci_aligned}
    \end{subfigure}
    \caption{\label{fig:uci_alignment}\textbf{Loss barriers for 250 weights across different optimization trajectories} (mean and standard deviations in parentheses) on the UCI Wisconsin Breast Cancer Diagnostics dataset. Aligning all weights to a single reference significantly reduces the loss barriers between the weights. } 
\end{figure}

\section{Classification with a Small Model} \label{sec:uci_classification}

\begin{table}[t!]
    \centering
    \begin{tabular}{lll}
        \toprule
        \textbf{Flow}  & \textbf{Accuracy} & \textbf{Loss} \\
        \midrule
        Euclidean                   & 0.998 $\pm$ 0.006     & 0.101 $\pm$ 0.050 \\ 
        Euclidean (aligned)         & 0.998 $\pm$ 0.006     & 0.070 $\pm$ 0.040 \\
        Euclidean (aligned + OT)    & 0.993 $\pm$ 0.010     & 0.053 $\pm$ 0.028 \\
        \midrule
        Normalized                  & 0.993 $\pm$ 0.009     & 0.027 $\pm$ 0.014 \\
        Normalized (aligned)        & 0.989 $\pm$ 0.011     & 0.030 $\pm$ 0.015 \\
        Normalized (aligned + OT)   & 0.988 $\pm$ 0.018	    & 0.044 $\pm$ 0.047 \\
        \midrule
        Geometric                   & 0.992 $\pm$ 0.011     & 0.019 $\pm$ 0.009 \\
        Geometric (aligned)         & 0.993 $\pm$ 0.001     & 0.018 $\pm$ 0.001 \\
        Geometric (aligned + OT)    & 0.991 $\pm$ 0.011 	& 0.020 $\pm$ 0.012 \\
        \midrule
        \textbf{Target}             & 0.992 $\pm$ 0.010     & 0.048 $\pm$ 0.032 \\
        \bottomrule
    \end{tabular}
    \caption{\label{tab:uci_class_table}\textbf{Test accuracy and loss ($\pm$ std) of the samples generated by flows with different design choices.} Euclidean and Normalized flows result in the most accurate weights, with Normalized flow generating lower-loss weights.}
\end{table}

After the toy Gaussian example, we move on to a relatively simple classification task to demonstrate that our flows can generate high-quality samples and compare the various design choices outlined in Chapter \ref{chapter:method}. The target model is an MLP with two hidden layers of 16 dimensions each and ReLU activations, on the binary classification task of the UCI Wisconsin Breast Cancer Diagnostic dataset \citep{streetNuclearFeatureExtraction1993}. We train all flows on the same dataset consisting of weights sampled from 100 independent trajectories optimized with Adam \citep{kingmaAdamMethodStochastic2017}. 

In addition to the availability of a large number of trained weights, this setup represents a preferable scenario in the sense that aligning all weights to a single reference eliminates the loss barriers between them to a large extent. Figure \ref{fig:uci_alignment} shows the pair-wise loss barriers (midpoint of the linear interpolation) before and after alignment for 250 randomly sampled weights from the set of Adam-optimized weights. The loss barriers are reduced considerably (from an average of 0.244 to 0.07) which supports the linear mode connectivity hypothesis for this setup. This should presumably make it easier to approximate the posterior distribution, as in the extreme case of zero-loss barriers between \textit{all} pairs of weights and assuming they capture all the modes, the posterior would be convex. 

\textbf{Sample Quality.} Table \ref{tab:uci_class_table} compares the predictive quality of the samples generated by different flows with or without alignment and mini-batch OT couplings. All flows can generate samples with almost perfect accuracy, matching or exceeding (although not significantly) the performance of Adam-optimized weights. Noticeably, the Geometric flow generates samples with lower loss than the Normalized flow, highlighting the benefit of modeling the vector field over the product geometry of normalized weights as well. Aligning all the weights to a reference before training, and training the flow with mini-batch OT couplings both decrease the loss of the generated samples for the Euclidean flow, but the same effect is not visible for the Normalized and Geometric flows. 

\begin{figure}[t!]
    \centering
    \includegraphics[width=\linewidth]{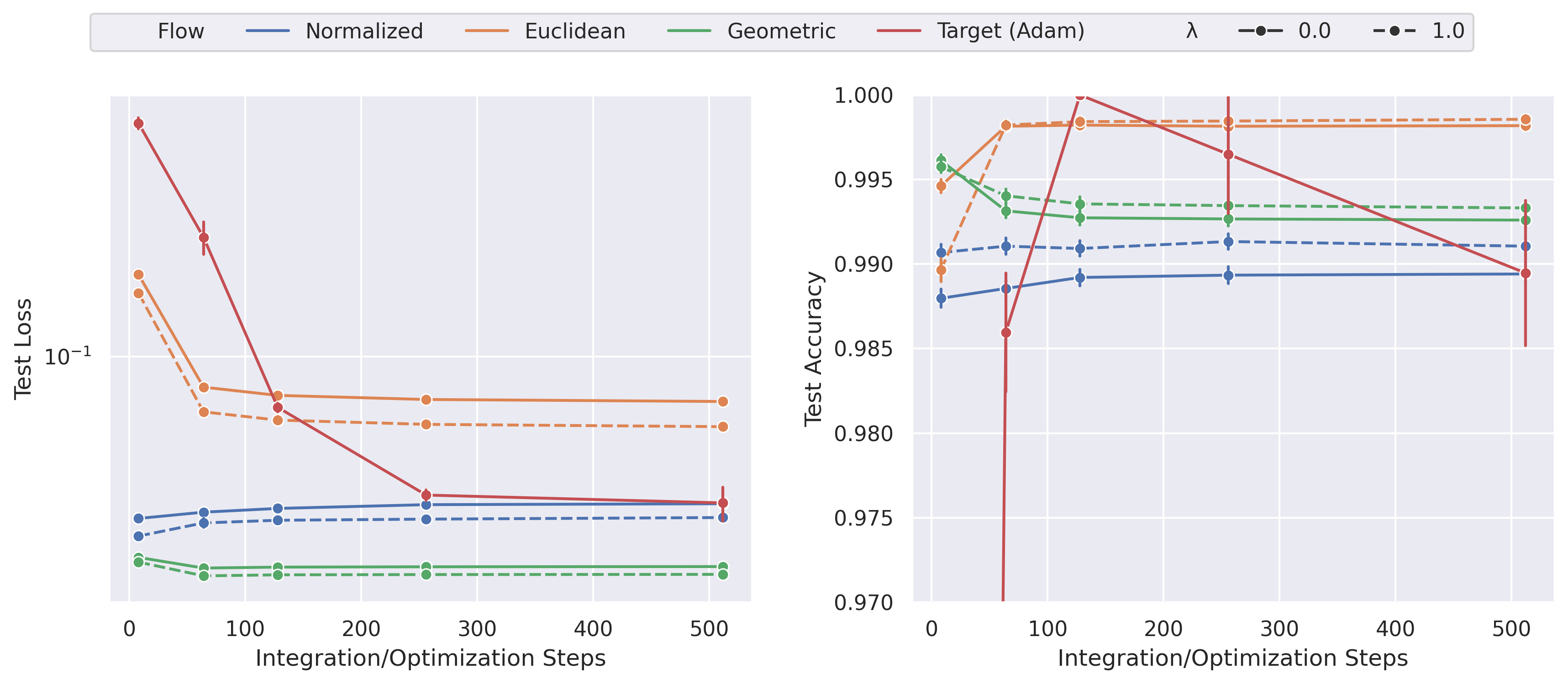}
    \caption{\label{fig:uci_steps}\textbf{Comparing the quality of the generated weights with Adam-optimized weights.} Weights generated by the Euclidean and Normalized flows have higher accuracy and lower loss than Adam-optimized weights, while the Geometric flow generated less accurate weights. Guidance during sampling improves sample quality.} 
\end{figure}

\textbf{Number of Integration Steps \& Guidance.} Figure \ref{fig:uci_steps} displays how the predictive performance of generated weights changes with the number of Euler steps performed to integrate the ODE. The main outcomes are:
\begin{itemize}
    \item Normalized and Geometric flows learn straighter paths compared to the Euclidean flow, apparent from the samples from the Euclidean flow showing a more significant improvement in quality going from 8 to 64 Euler steps. 
    \item For all three flows, guiding the integration with task gradients improves sample quality, but increasing the number of integration does not appear to significantly increase the effect of guidance. 
\end{itemize}

\section{MNIST Classification} \label{sec:mnist_classification}

We now move on to a harder classification task and larger models to evaluate different use cases of a weight-space flow. We start with the MNIST dataset, and an MLP with a single hidden layer of 10 neurons and ReLU activations, following the setup used in \citep{peeblesLearningLearnGenerative2022}, but we sample a smaller dataset over 20 independent optimization trajectories.

This corresponds to a harder task than the UCI classification of the previous section for several reasons. The base MLP has around an order-of-magnitude of more parameters (7,960 rather than 802 parameters) while our GNN is in fact smaller. Moreover, the optimized weights do not reach perfect accuracy unlike in the UCI classification task. 

\begin{table}[t!]
    \centering
    \begin{tabular}{llll}
        \toprule
        \textbf{Flow}  & \textbf{Accuracy} & \textbf{Loss} & \textbf{Diversity} \\
        \midrule 
        Euclidean (aligned)              & 0.737 $\pm$ 0.085	& 0.814 $\pm$ 0.286 & 0.00102 $\pm$ 0.00114 \\
        Euclidean (aligned + OT)          & 0.757 $\pm$ 0.077 & 0.753 $\pm$ 0.245 & 0.00085 $\pm$ 0.00094 \\
        \midrule
        Normalized (aligned)          & 0.753 $\pm$ 0.074	& 1.537 $\pm$ 0.046 & 0.00086 $\pm$ 0.00055 \\
        Normalized (aligned + OT)         & 0.706 $\pm$ 0.078 & 1.608 $\pm$ 0.047 & 0.00117 $\pm$ 0.00074 \\
        \midrule
        Geometric (aligned)               & 0.737 $\pm$ 0.070	& 1.457 $\pm$ 0.040 & 0.00073 $\pm$ 0.00049 \\
        Geometric (aligned + OT)     & 0.786 $\pm$ 0.064 & 1.443 $\pm$ 0.046 & 0.00093 $\pm$ 0.00063 \\
        \midrule
        \textbf{Target}         & 0.933 $\pm$ 0.009 & 0.231 $\pm$ 0.027 & 0.00002 $\pm$ 0.00001  \\
        \bottomrule
    \end{tabular}
    \caption{\label{tab:mnist_class_table}\textbf{Test accuracy, loss ($\pm$ std), and diversity of the samples generated by flows for classification on the MNIST dataset after alignment.} Geometric flow with OT couplings results in the best-performing samples, but most flows are within one standard deviation of each other. }
\end{table}

\begin{figure}[t!]
    \centering
    \includegraphics[width=\linewidth]{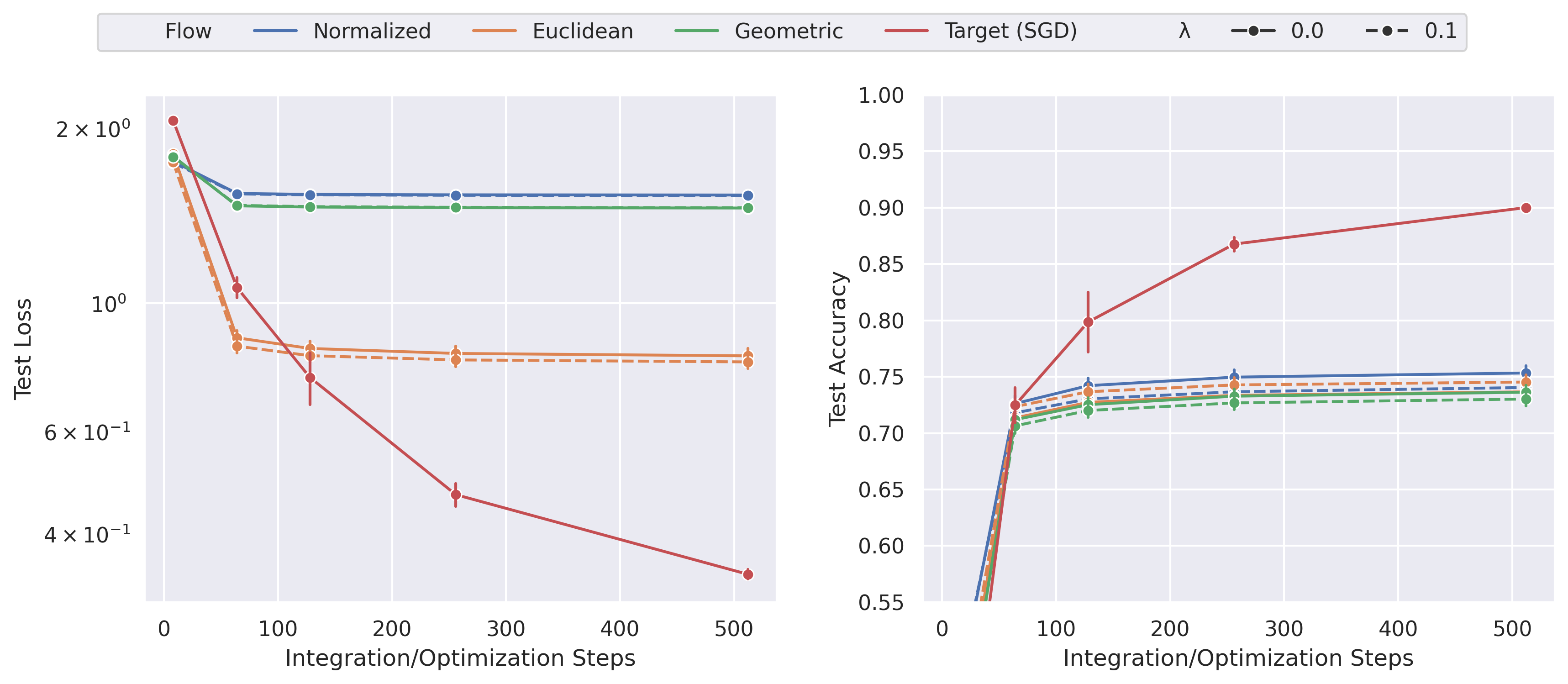}
    \caption{\label{fig:mnist_steps}\textbf{Comparing the quality of the generated weights with SGD-optimized weights.} After a very small number of steps, sampled weights are comparable in performance with SGD-optimized weights but plateau after around 64 steps. Geometric and Normalized flows again appear to lead to straighter paths.} 
\end{figure}

\subsection{Sample Quality and Diversity} \label{sec:sample_quality_diversity}

Table \ref{tab:mnist_class_table} shows the predictive performance and functional diversity of the weights generated by different kinds of flows trained on aligned weights. Functional diversity is a desirable property of a weight-space generative model, as more diverse weights could correct for each other's errors and make model averaging more effective. We focus on functional rather than parametric diversity since even if two neural networks do not have identical parameters, they could be computing the same function, and thus provide no benefit for model averaging. To measure the functional diversity in a set of neural networks, we compute the average pairwise Jensen-Shannon divergence (JSD) \citep{endresNewMetricProbability2003,mishtalJensenShannonDivergenceEnsembles2012} between the output (class probabilities) distributions of neural networks in the set. For a set of weights $S = \{ \theta_i \}_{i=1}^N$ and model $f$, we fix input points $\{ x_i \}_{i=1}^K$ and compute
\begin{equation} \label{eq:diversity}
    \text{Diversity}(S) := \frac{1}{N^2} \sum_{i, j} \text{JSD}\left(
        \{ f(\theta_i, x_l)\}_{l=1}^K, 
        \{f(\theta_j, x_l)\}_{l=1}^K
    \right)
\end{equation}
with JSD defined as 
\begin{equation}
    \text{JSD}(A, B) := 
    \frac{1}{2} \sum_{x \in A} p_A(x) \log \left( \frac{2p_A(x)}{p_A(x) + p_B(x)} \right) + 
    \frac{1}{2} \sum_{x \in B} p_B(x) \log \left( \frac{2p_B(x)}{p_A(x) + p_B(x)} \right)
\end{equation}
and we fit kernel density estimators to both sets to estimate the densities $p_A$ and $p_B$. 

Table \ref{tab:mnist_class_table} displays the test accuracy, loss, and diversity (Equation \ref{eq:diversity}) of the samples generated with each flow, trained after alignment, and with independent as well as mini-batch OT couplings. While the accuracy of sampled weights are generally within one standard deviation of each other, the Geometric flow with OT couplings results in the highest accuracy. However, unlike the previous UCI classification task, the sampled weights individually perform worse than optimized weights. 

The output distributions of the generated weights shows higher diversity than SGD-optimized weights, although this is also influenced by the generated weights performing worse. Nevertheless, as we will observe in the next chapter, this diversity makes model averaging more effective and considerably increases the predictive performance of generated weights. 

Figure \ref{fig:mnist_steps} shows how sample quality varies with the number of steps for our three flows trained with independent couplings, and sampled with and without guidance. Unlike the previous task where the sampled weights were directly of almost perfect accuracy, the sample quality for MNIST flows plateau after a certain number of steps, and SGD-optimized weights show superior performance. Similar to the results on the UCI classification task, Normalized and Geometric flows also show less improvement with a larger number of steps, indicating that the flows have learned straighter paths then the Euclidean flow. 

\subsection{Posterior Predictive Performance} \label{sec:posterior_predictive}

\begin{figure}[t!]
    \centering
    \begin{subfigure}{0.55\linewidth}
        \centering
        \includegraphics[width=\linewidth]{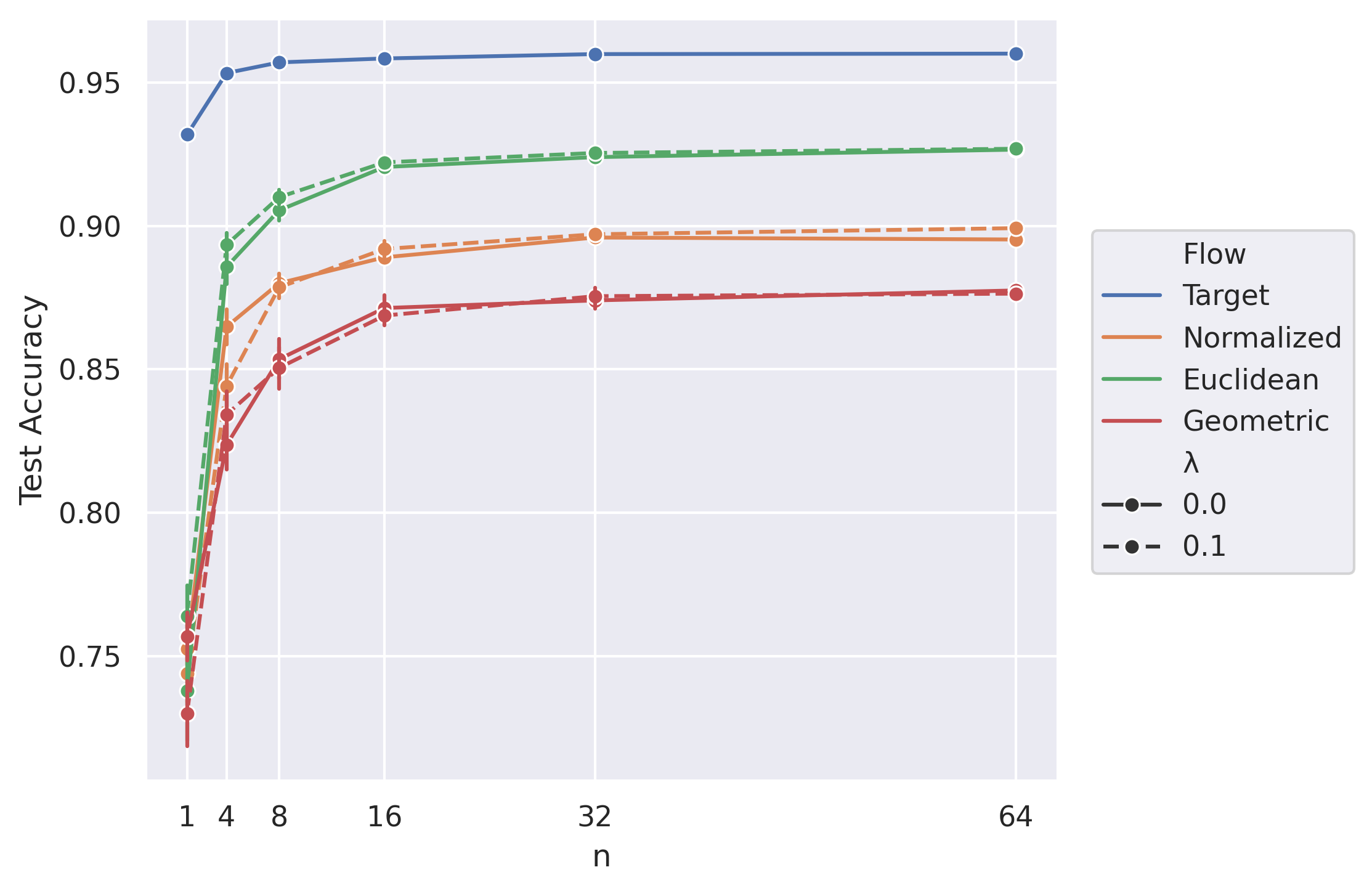}
        \caption{Independent Couplings}
    \end{subfigure}
    \begin{subfigure}{0.43\linewidth}
        \centering
        \includegraphics[width=\linewidth]{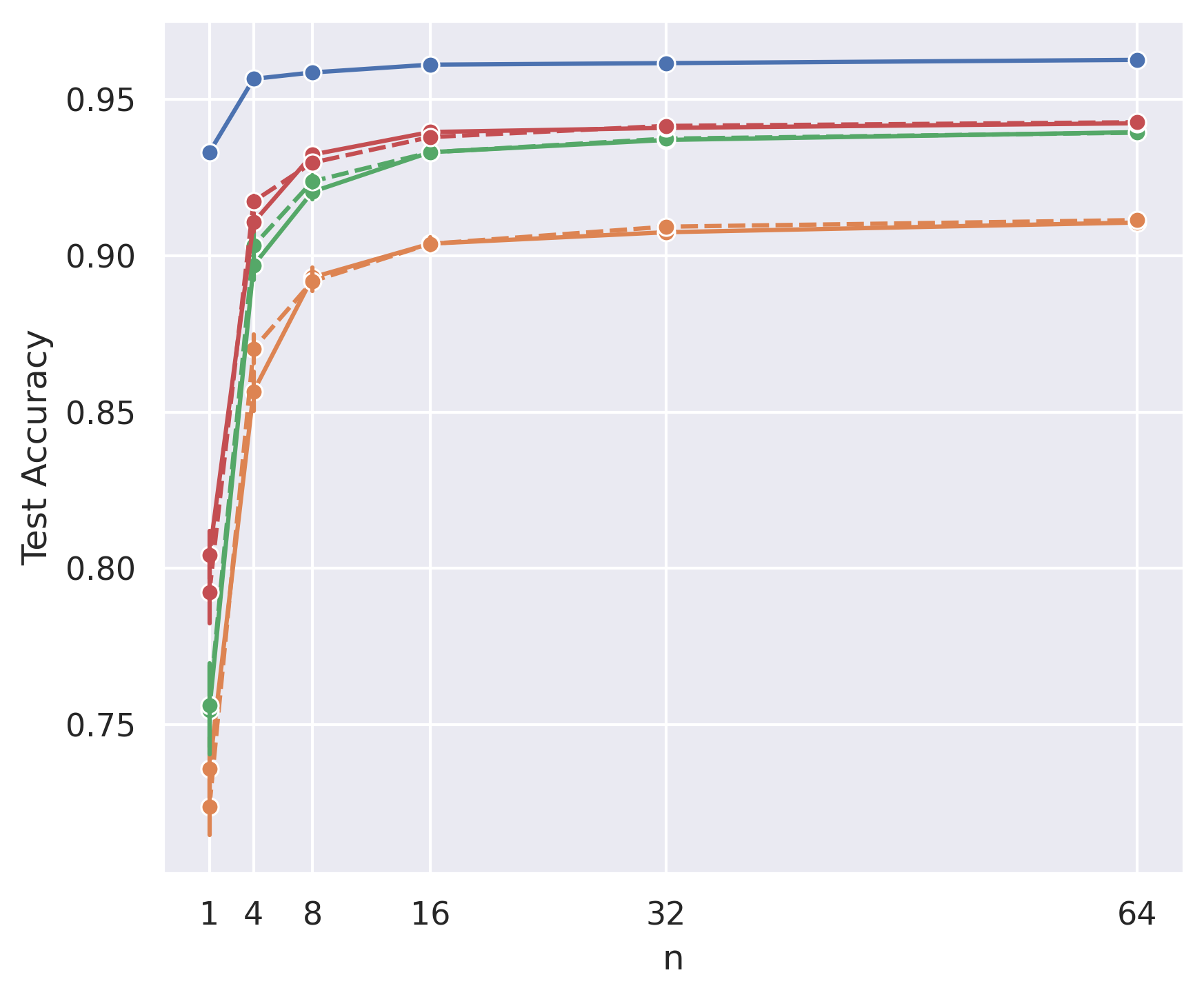}
        \caption{OT Couplings}
    \end{subfigure}
    \caption{\label{fig:mnist_averaging}\textbf{Accuracy of the predictions averaged over various numbers of samples from the flows' push-forward distributions, comparing independent and OT couplings.} The Euclidean flow shows the highest accuracy, reaching the accuracy of individual SGD-optimized weights.} 
\end{figure}

We now evaluate the predictive performance of the sampled weights after Bayesian model averaging; i.e. for each flow, we sample from its push-forward distribution $\hat p_1 (\theta)$ and compute a Monte Carlo estimate for the expectation $\bbE_{\hat p_1 (\theta)} \left[ p(y\vert x, \theta) \right]$, as in Equation \ref{eq:model_averaging}. Figure \ref{fig:mnist_averaging} displays the accuracy of predictions averaged over increasing numbers of samples from $\hat p_1$, comparing our three flows as well as independent and mini-batch OT couplings. With independent couplings, the Euclidean flow results in the highest accuracies, noticeably reaching the accuracy of individual SGD-optimized weights. 

Training the flows with OT couplings improves the predictive performance of all three after model averaging. The improvement for the Geometric flow is more significant than that of the Euclidean and Normalized flows, moving from $\sim87\%$ to $~94\%$. This supports the results in Table \ref{tab:mnist_class_table} that the Geometric flow benefits the most from OT couplings. 

\subsection{Transferability to Different Tasks} \label{sec:task_generalization}

\begin{table}[t!]
    \centering
    \begin{tabular}{lll}
        \toprule
        \textbf{Flow} & \textbf{Accuracy} & \textbf{Loss} \\
        \midrule
        \textbf{Adam-optimized} & 0.908 $\pm$ 0.003 & 0.334 $\pm$ 0.008 \\
        \midrule
        \textbf{With Guidance} & & \\
        Euclidean   & 0.754 $\pm$ 0.082     & 0.764 $\pm$ 0.252 \\
        Normalized  & 0.724 $\pm$ 0.081     & 1.543 $\pm$ 0.045 \\
        Geometric   & 0.730 $\pm$ 0.067     & 1.470 $\pm$ 0.043 \\
        \midrule
        \textbf{No guidance} & 0.080 $\pm$ 0.030 & 9.601 $\pm$ 1.402 \\ 
        \bottomrule
    \end{tabular}
    \caption{\label{tab:mnist_generalization}\textbf{Predictive performance of samples from the MNIST flows with guidance ($\lambda = 0.1$) from Fashion-MNIST task gradients.} Although the base model (Adam) accuracy is lower than for MNIST, samples obtained with guidance reach an accuracy similar to that of the samples obtained without guidance on MNIST, indicating that a flow trained on one dataset can be transferred to another dataset via guidance. }
\end{table}

As a potential use case of our flows, we now evaluate the effectiveness of a flow trained on weights from a dataset (e.g. MNIST) different than the target dataset (e.g. Fashion-MNIST) using the same architecture, although the GNNs we model our flow with can be applied to architectures other than the one they were trained on. There are three main ways we can evaluate this performance:
\begin{enumerate}
    \item Evaluate the samples directly.
    \item Guide sampling with gradients from the target task. 
    \item Initialize model with sampled weights and train on the target dataset.
\end{enumerate}

First, Table \ref{tab:mnist_generalization} shows the performance (on Fashion-MNIST) of samples from flows trained with MNIST weights but sampled with guidance from the Fashion-MNIST task gradients over 512 Euler steps. First, expectedly, the last row of the table shows that samples obtained directly from the MNIST flows without guidance achieve an average accuracy of 8\%, which is not better than random guessing. However, although the accuracy of the base model after optimization with Adam is slightly lower for Fashion-MNIST compared to MNIST, samples from all three flows achieve a test accuracy of at least 72\%, which is similar performance of the samples obtained without guidance from the same flows on MNIST. 

These results highlight that even though the flows are trained on weights obtained over one dataset, they can capture some salient features shared across optimization landscapes, and can be adapted to different datasets without having to train them from scratch, in this case only by guiding the sampling process with task gradients which is significantly cheaper than training the entire flow from scratch.  

\begin{figure}[t!]
    \centering
    \begin{subfigure}{0.47\linewidth}
        \centering
        \includegraphics[width=\linewidth]{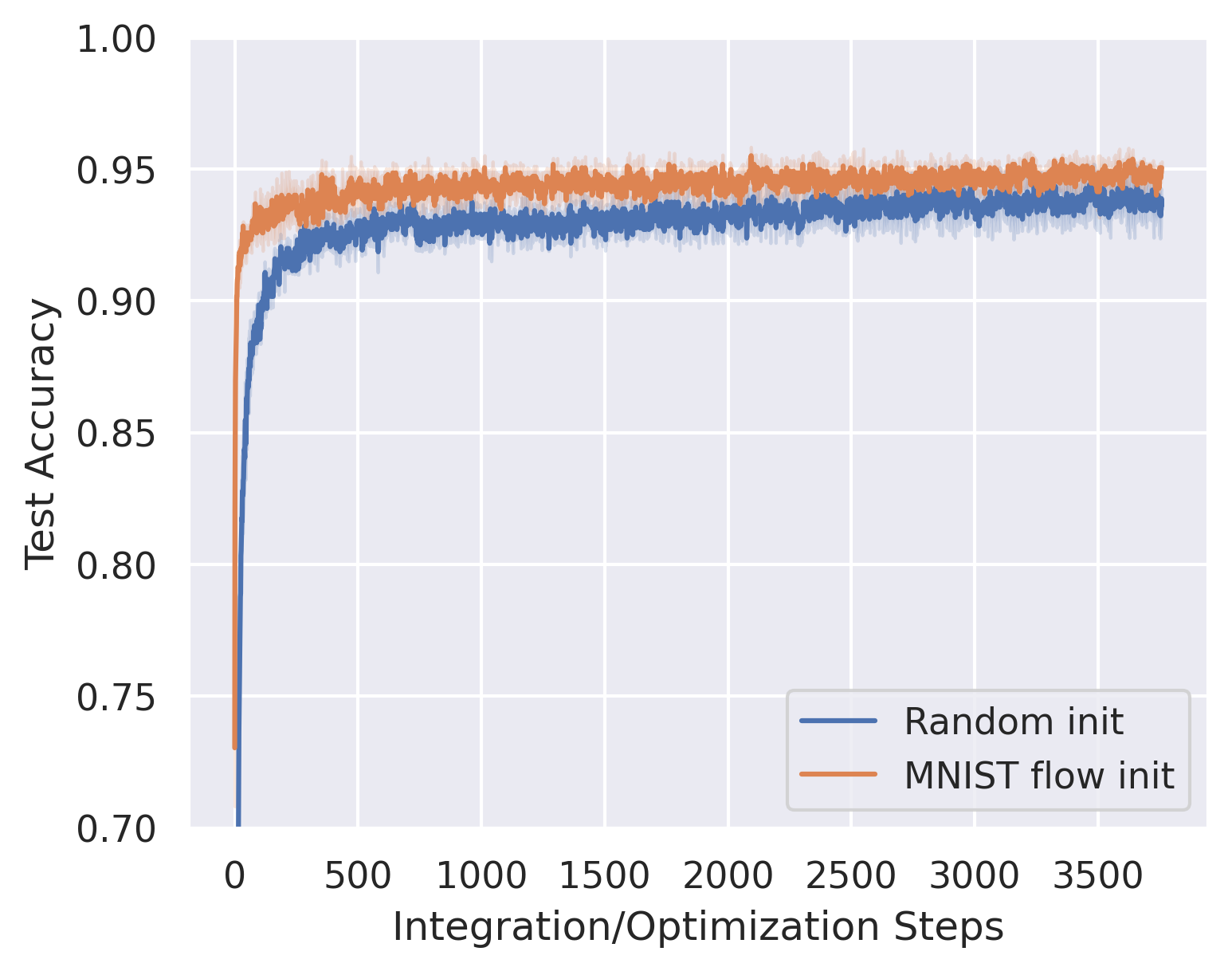}
        \caption{Test Accuracy}
    \end{subfigure}
    \begin{subfigure}{0.47\linewidth}
        \centering
        \includegraphics[width=\linewidth]{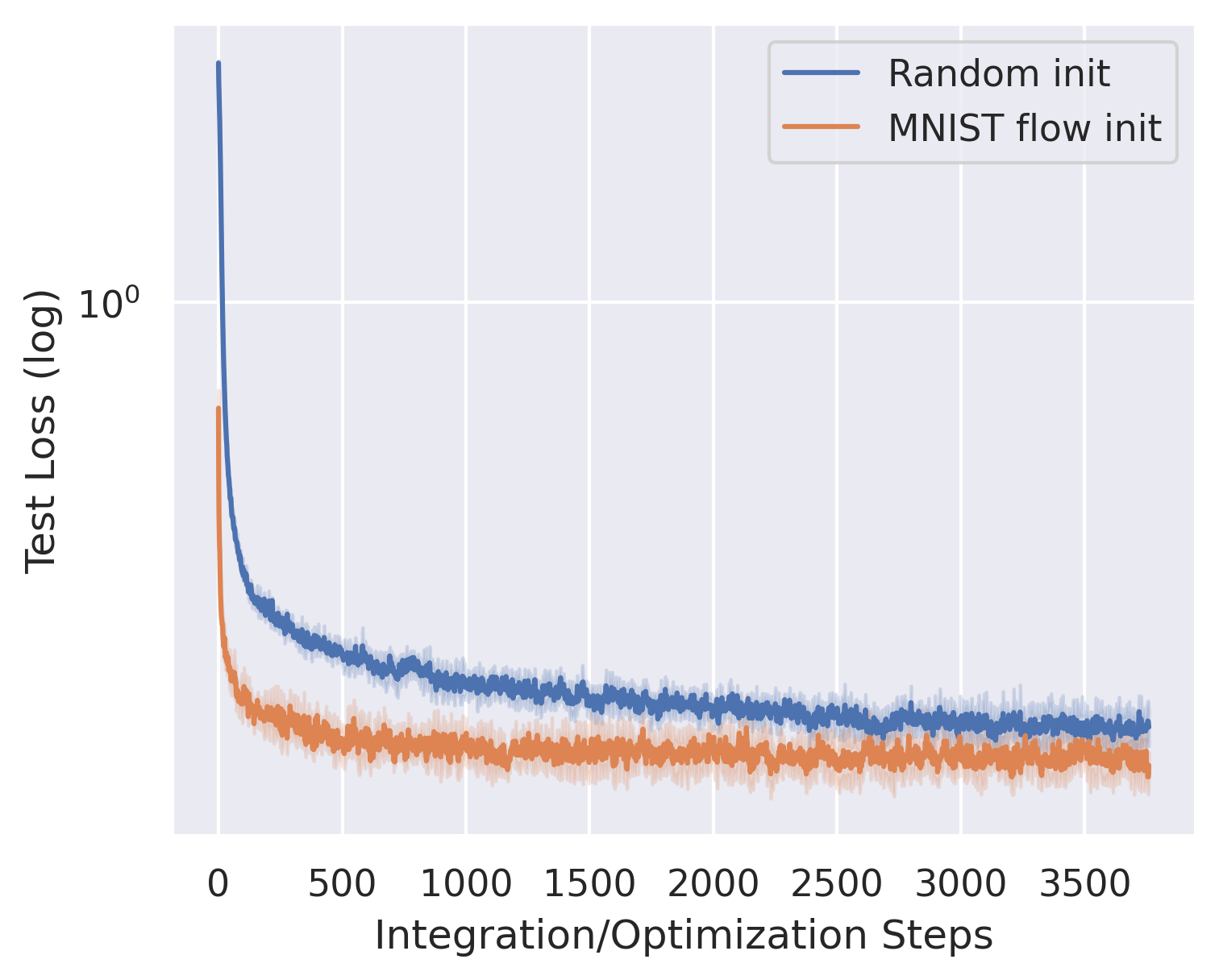}
        \caption{Test Loss}
    \end{subfigure}
    \caption{\label{fig:mnist_init}\textbf{Optimization trajectories of Adam on Fashion-MNIST with initial weights sampled from an isotropic Gaussian and the Euclidean flow trained on MNIST weights.} While the curves tend to converge towards similar values, the learned initialization converges faster.} 
\end{figure}

Next, we evaluate if the samples from the MNIST flows obtained without any guidance provide a more effective initialization scheme then random initialization for a model trained on Fashion-MNIST. Figure \ref{fig:mnist_init} displays the test accuracy and loss values during training when the model is initialized from an isotropic Gaussian or with our flow trained on MNIST weights, averaged over five training runs. Although the curves tend to converge towards similar values, the model initialized with the weights from the flow converges faster. This indicates that using our flows for learned initialization on similar tasks can lead to faster convergence and thus more efficient training.

\begin{table}[t!]
    \centering
    \begin{tabular}{lll}
        \toprule
        \textbf{Flow} & \textbf{Accuracy} & \textbf{Loss} \\
        \midrule
        Euclidean                   & 0.826 $\pm$ 0.076     & 0.830 $\pm$ 0.281 \\
        Euclidean (w/ guidance)     & 0.842 $\pm$ 0.079     & 0.796 $\pm$ 0.337 \\
        \midrule
        Geometric                  & 0.890 $\pm$ 0.030     & 1.030 $\pm$ 0.038 \\
        Geometric (w/ guidance)    & 0.886 $\pm$ 0.033     & 1.031 $\pm$ 0.037 \\
        \midrule
        Normalized                   & 0.203 $\pm$ 0.122     & 2.652 $\pm$ 0.667 \\
        Normalized (w/ guidance)     & 0.184 $\pm$ 0.083     & 2.776 $\pm$ 0.575 \\
        \bottomrule
    \end{tabular}
    \caption{\label{tab:arch_mnist_generalization}\textbf{Generalization of flows trained on MNIST MLPs to MLPs with different architectures.} Euclidean and Geometric flows can generalize to an architecture with a wider hidden layer without requiring guidance while the Normalized flow fails to do so even with guidance.}
\end{table}

\subsection{Transferability to Different Architectures} \label{sec:arch_generalization}

Since a GNN is by design not limited to a certain graph structure, our weight-space flows can also be applied to architectures different than the one they were trained on. To evaluate this, we use the flow models from the previous chapters trained on the weights of an MLP with 10 hidden neurons and evaluate them using a larger MLP with 32 hidden neurons. Table \ref{tab:arch_mnist_generalization} shows the resulting test accuracies and losses on MNIST. In particular the Euclidean and Geometric flows output samples with even higher accuracy than they did for the smaller MLP. This is not surprising as the larger model is expected to have better performance than the smaller model, but shows that the flows learn to approximate the posterior for not just one architecture, but for the underlying task more generally. 

\subsection{Model Size Scaling} \label{sec:scaling}

\begin{figure}[t!]
    \centering
    \begin{subfigure}{0.47\linewidth}
        \centering
        \includegraphics[width=\linewidth]{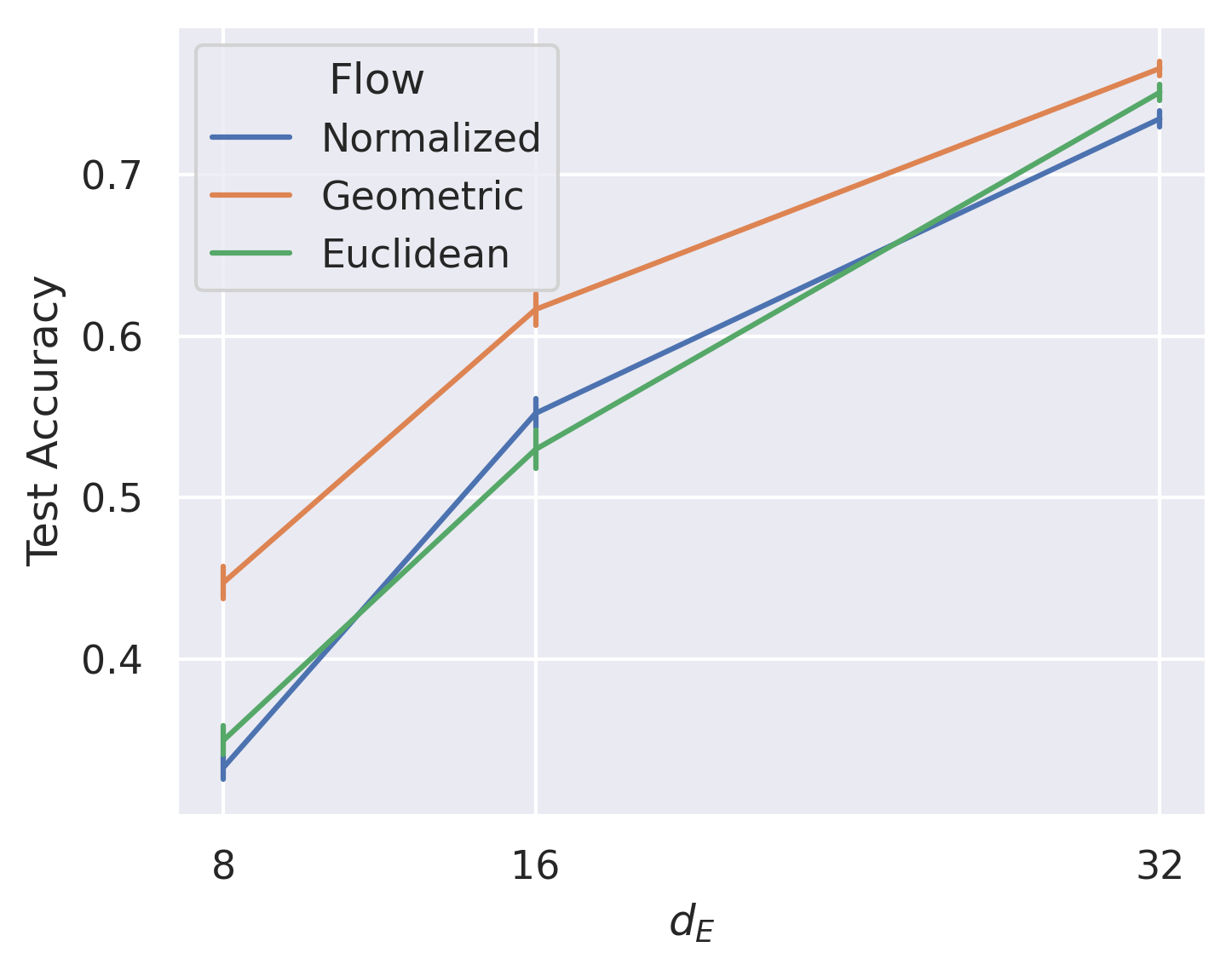}
        \caption{\label{scale_acc}Test Accuracy}
    \end{subfigure}
    \begin{subfigure}{0.47\linewidth}
        \centering
        \includegraphics[width=\linewidth]{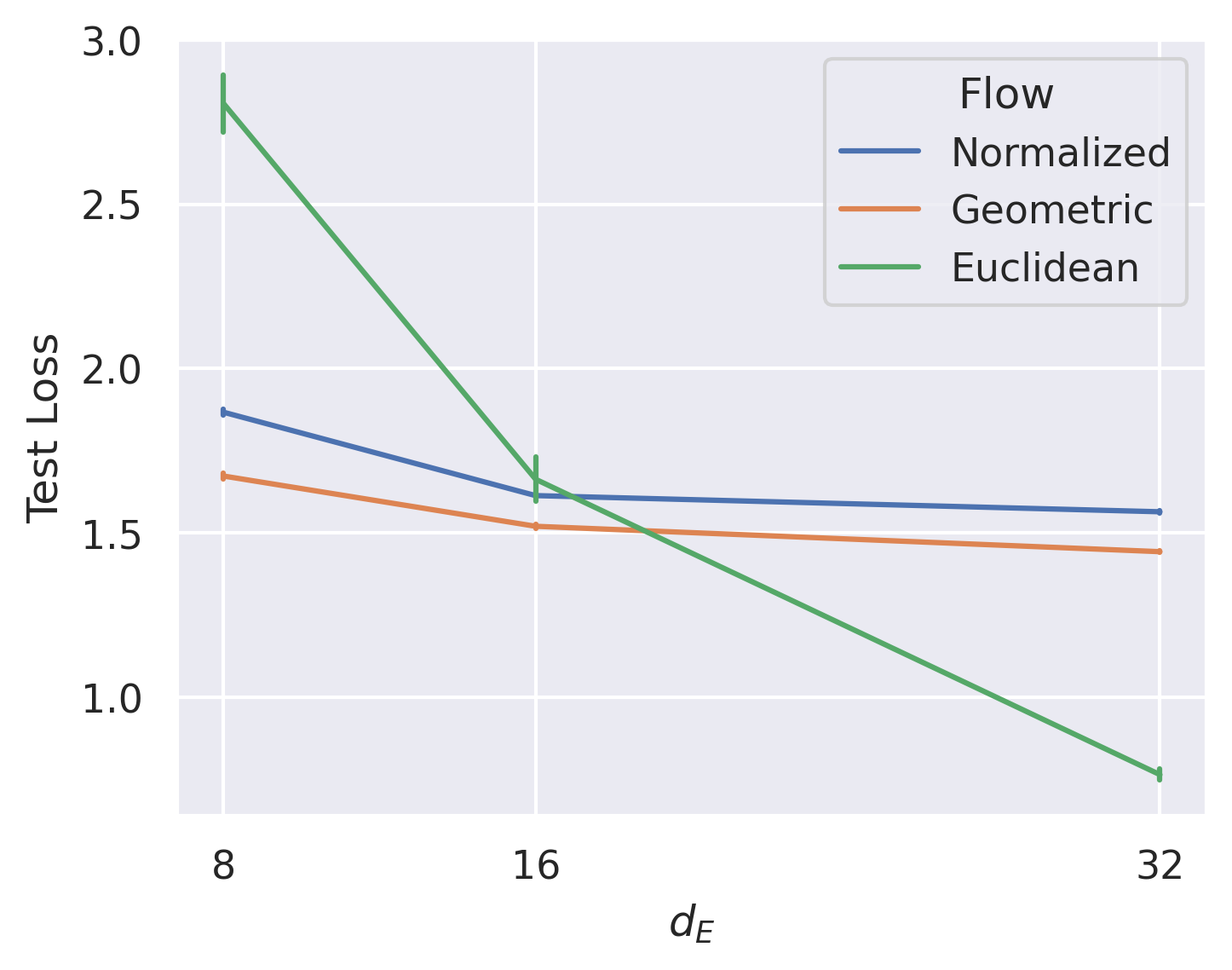}
        \caption{\label{scale_loss}Test Loss}
    \end{subfigure}
    \caption{\label{fig:scaling}\textbf{Sample predictive performance with increasing model size}, characterized by the node/edge feature dimension $d_E$. The upward trajectory in Figure \ref{scale_acc} and the downwards trajectory in Figure \ref{scale_loss} implies that there are still performance gains to be achieved by using larger models.} 
\end{figure}

We conclude our evaluations by measuring if there is performance gains to be expected by using larger flow models than the ones we have considered so far. We thus train flow models with different node/edge feature dimensions ($d_E$) for the same number of iterations, and evaluate their sample qualities. The largest model, with 32 dimensional features, has approximately 146K parameters while the smallest model with 8 dimensional features has approximately 15K parameters. 

Figure \ref{fig:scaling} displays the quality of the generated samples for flow models with 8, 16, and 32 dimensional features. While the Geometric flow achieves the most accurate samples for each model size, the upwards trajectories for the test accuracies and the downwards trajectories for the test losses common to all three kinds of flows indicate that there are performance gains to be expected by using larger models. 

Considering that the literature on weight-space generative models often uses models with number of parameters on the order of tens or even hundreds of millions, the largest models in Figure \ref{fig:scaling} are relatively small, and still manage to achieve high sample quality.

% !TeX root = ../main.tex
% Add the above to each chapter to make compiling the PDF easier in some editors.

\chapter{Conclusion} \label{chapter:discussion}

Deep generative models, applied to domains such as images \citep{esserScalingRectifiedFlow2024b} and molecules \citep{abramsonAccurateStructurePrediction2024} to great effect, often take into account various symmetries in their data such as translation-invariance in images or rotational symmetries for molecules. Generative models applied to other neural networks' weights on the other hand, have so far only considered the permutation symmetries of neural networks, and not the scaling symmetries arising from the use of non-linear activations such as ReLU. We have thus attempted to address this gap, and showed that utilizing this geometric structure of neural networks in addition to the permutation symmetries helps build more effective generative models of neural network weights. 

Using the flow matching framework, we have constructed three different kinds of flows with different ways of handling this geometry, and evaluated them in various use cases such as transfer learning, learned initialization, and Bayesian model averaging. Our results have demonstrated that often the fully geometric flow results in better performance, and that there are still gains to be expected by further scaling up our models. 

\section{Future Directions}

While we have demonstrated that generative models in weight-space trained via the flow matching objective can generate high-quality samples and that taking into account the geometry of the neural network weights can make such flows more effective, applying modern generative models to neural network weights is still an active research area with many fruitful potential future directions and we outline a list of potential directions for future work in this section. 

\subsection{More Fine-Grained Geometric Considerations}

Our flows so far are built for MLPs and only take into account the permutation symmetries between subsequent layers and scaling symmetries resulting from the use of ReLU activations. However, as outlined in previous work \citep{kofinasGraphNeuralNetworks2024,limGraphMetanetworksProcessing2023}, different architectural choices such as convolutions, residual connections, or transformer blocks induce different kinds of symmetries that can be captured by constructing the neural graphs accordingly. The same GNN architectures, such as the Relational Transformer we have used, can then be used for learning tasks over these graphs. 

Different activation functions also induce different symmetries \citep{godfreySymmetriesDeepLearning2022} which can be used to embed neural networks into different manifolds, similar to our embedding of ReLU MLPs on a product of hyperspheres and Euclidean space following \citep{pittorinoDeepNetworksToroids2022}. Since generative modeling frameworks such as flow matching can readily be applied to different manifolds \citep{chenRiemannianFlowMatching2023}, including the use of data-dependent metrics \citep{kapusniakMetricFlowMatching2024} rather than explicitly formulating a manifold, extending our flow models to different kinds of symmetries represents another potentially valuable line of work. 

Another future direction related to symmetries in weight-space is to account for data-dependent symmetries \citep{zhaoSymmetriesFlatMinima2023} that arise from the data the model is trained on rather than the static symmetries in the previous paragraph that are valid for all instantiations of the same architecture. Recent work attempting to find these symmetries in an automated way \citep{zhaoFindingSymmetryNeural2024} can also be useful building blocks in this light. 

Finally, certain functional constraints in neural networks such as group equivariance are often imposed through constraints on the neural networks' weights \citep{weilerEquivariantCoordinateIndependent2023}. If the base model has such constraints, accounting for them while designing the flow as well could reduce the effective dimensionality of the problem considerably and lead to more efficient flows. 

\subsection{Generative Modeling and Training}

While each our flows is trained on a single architecture and dataset to push-forward a single source distribution to a single target distribution, a single flow that could potentially learn to map different source distributions (e.g. corresponding to the same base architecture on different datasets, or different architectures on the same dataset) could be a more directly useful method. This would essentially correspond to a setting where the model learns to model a vector field not in weight-space, but in the space of probability distributions. Generalizations of the flow matching framework to this setting, such as Meta Flow Matching \citep{atanackovicMetaFlowMatching2024a} and Wasserstein Flow Matching \citep{havivWassersteinFlowMatching2024} could be useful building blocks for such a flow, as well as the publicly available datasets of neural network weights \citep{schurholtModelZoosDataset2022}. This approach could lead to weight-space ``foundation models'' that are trained over a diverse set of tasks and can be adapted to specific use cases by fine-tuning on a single task. While this would be instance of meta-learning \citep{hospedalesMetaLearningNeuralNetworks2022}, it would be have the added benefit of obtaining probability distributions over solutions rather than point estimates. 

Furthermore, the requirement of sample-based training can also be relaxed by utilizing recent work on generative modeling without samples \citep{vargasTransportMeetsVariational2023,akhound-sadeghIteratedDenoisingEnergy2024}. In particular, iDEM \citep{akhound-sadeghIteratedDenoisingEnergy2024} learns a sampler with access to the energy function and its gradient (and optionally samples from the posterior) without having to simulate the forward and backward trajectories of the sampling process, making it potentially useful in high-dimensional spaces such as neural network weights. 

\subsection{Sampling and Guidance}

We sample from our flows using an Euler ODE solver and perform guidance during sampling only using the base task gradients, but the sampling phase has a richer design space that can be utilized to obtain faster and more controllable flows. To begin with, higher-order ODE solvers can be used to reduce the approximation error to the learned vector field at the cost of longer sampling times. Distillation methods such as Flow Map Matching \citep{boffiFlowMapMatching2024} and Consistency Flow Matching \citep{yangConsistencyFlowMatching2024} can be utilized to instead speed up the sampling process.  

Using a generative modeling framework such as flow matching also enables guidance methods beyond using task gradients. First, any differentiable objective can be used in place of the task loss to guide the sampling process. Performing classifier-free guidance is also possible for flow models \citep{zhengGuidedFlowsGenerative2023}, to instead condition the sampling process on a variable such as the desired loss, as also done in weight-space with diffusion models in \citet{peeblesLearningLearnGenerative2022}. Finally an alternative conditioning approach in flow matching is to differentiate through the ODE sampling process to optimize the initial point $x_0$ so that the solution $x_1$ minimizes a loss function \citep{ben-hamuDFlowDifferentiatingFlows2024}. This and future work for controllable sampling in flow models can be adapted neural network weights to condition flows on desired quantities or objectives such as adversarial robustness.

\appendix{}

% !TeX root = ../main.tex
% Add the above to each chapter to make compiling the PDF easier in some editors.

\chapter{Experimental Setups}\label{appendix:experimental_setups}

\section*{Implementation Details and Computational Resources}

Our flows were primarily implemented in PyTorch \citep{paszkePyTorchImperativeStyle2019a}, building on the weight-space GNN and graph construction implementation of \citep{kofinasGraphNeuralNetworks2024} and with our custom flow matching implementation utilizing parts of the \texttt{torchcfm} package of \citep{tongImprovingGeneralizingFlowbased2023}. We have implemented a custom Euler solver to integrate our ODEs and used the \texttt{geoopt} package \citep{kochurovGeooptRiemannianOptimization2020} for geometric operations.

Throughout the experiments, we train all our models using a combination of NVIDIA A100 an H100 GPUs, with each model being trained on a single GPU. The training times range from around 6 hours for the smaller Euclidean flows to approximately 40 hours for the larger Geometric flows, although our implementation was not fully optimized for efficiency. 

\section*{Flow Between Gaussians (Section \ref{sec:gaussian_flow})}

\subsection*{Data}

\begin{itemize}
    \item $p_0 = \N(0, \mathbf{I}), p_1 = \N(1, \mathbf{I})$. 
    \item The model is an MLP with dimensions 30 - 16 - 16 - 2 and ReLU activations. 
\end{itemize}

\subsection*{Flow Architecture}
\begin{itemize}
    \item Relational Transformer \citep{diaoRelationalAttentionGeneralizing2023,kofinasGraphNeuralNetworks2024} with 5 layers and $d_E = 32$, GeLU activations \citep{hendrycksGaussianErrorLinear2023a}, one attention head. 
\end{itemize}

\subsection*{Flow/Training}
\begin{itemize}
    \item Independent coupling, $t \sim \text{Beta}(1, 2)$, Gaussian probability path with $\sigma = 0.001$, $x_1$ prediction.
    \item Trained for 20,000 iterations with batch size 16, using the Adam optimizer with initial learning rate 0.001.
\end{itemize}

\section*{UCI Classification (Section \ref{sec:uci_classification})}

\subsection*{Data}

\begin{itemize}
    \item $p_0 = \N(0, 0.1 \mathbf{I})$. To obtain samples from $p_1$, we sample 100 models from $p_0$ and train each independently for 50 epochs with Adam. We ignore the first 10 epochs, and record one sample every four iterations after that. 
    \item We use the same MLP architecture as the Gaussian experiments. 
\end{itemize}

\subsection*{Flow Architecture}
\begin{itemize}
    \item Relational Transformer \citep{diaoRelationalAttentionGeneralizing2023,kofinasGraphNeuralNetworks2024} with 5 layers and $d_E = 64$, GeLU activations \citep{hendrycksGaussianErrorLinear2023a}, one attention head. 
\end{itemize}

\subsection*{Flow/Training}
\begin{itemize}
    \item $t \sim \text{Beta}(1, 2)$, Gaussian probability path with $\sigma = 0.0001$, $x_1$ prediction.
    \item Trained for 250,000 iterations with batch size 32, using the Adam optimizer with initial learning rate 0.001.
\end{itemize}

\section*{MNIST Classification (Section \ref{sec:mnist_classification})}

\subsection*{Data}

\begin{itemize}
    \item $p_0 = \N(0, 0.1 \mathbf{I})$. For the target samples, we independently train 50 models for 25 epochs using SGD with momentum 0.9, learning rate 0.1, and weight decay 0.00001. We collect one sample every 10 iterations, discarding the first 5 epochs. 
    \item The base model is an MLP with 784 input and 10 output dimensions, and one hidden layer of 10 units. 
\end{itemize}

\subsection*{Flow Architecture}
\begin{itemize}
    \item Relational Transformer \citep{diaoRelationalAttentionGeneralizing2023,kofinasGraphNeuralNetworks2024} with 5 layers and $d_E = 32$, GeLU activations \citep{hendrycksGaussianErrorLinear2023a}, one attention head. 
\end{itemize}

\subsection*{Flow/Training}
\begin{itemize}
    \item $t \sim \text{Beta}(1, 2)$, Gaussian probability path with $\sigma = 0.00001$, $x_1$ prediction.
    \item Trained for $100,000$ iterations with batch size 8. 
\end{itemize}

\microtypesetup{protrusion=false}

% \addchap{Abbreviations}
% \begin{acronym}
% 	\itemsep-.25\baselineskip
% 	\acro{TUM}[TUM]{Technical University of Munich}
% 	% TODO: add acronyms
% \end{acronym}

\listoffigures{}
\listoftables{}
\microtypesetup{protrusion=true}
\printbibliography

\end{document}